 \documentclass[final,5p,times,twocolumn]{elsarticle}

\usepackage{graphicx}
\usepackage{amssymb}

\usepackage{lineno}

\usepackage{tikz}
\usepackage{xcolor}
\usepackage{color}
\usepackage{natbib}
\usepackage{url}
\usepackage{subcaption}
\usepackage{makecell} 
\usepackage{comment}

\journal{Journal of Photogrammetry and Remote Sensing}

\begin{document}

\begin{frontmatter}


\title{Counting of Grapevine Berries in Images via Semantic Segmentation using Convolutional Neural Networks}



\author[igg]{Laura Zabawa}
\author[jki]{Anna Kicherer}
\author[igg]{Lasse Klingbeil}
\author[jki]{Reinhard T\"opfer}
\author[igg]{Heiner Kuhlmann}
\author[rs]{Ribana Roscher}

\address[igg]{Bonn University, Department of Geodesy, Institute for Geodesy and Geoinformation}
\address[rs]{Bonn University, Remote Sensing Group, Institute for Geodesy and Geoinformation}
\address[jki]{Julius K\"uhn-Institut, Federal Research Centre of Cultivated Plants, Institute for Grapevine Breeding Geilweilerhof}

\begin{abstract}
The extraction of phenotypic traits is often very time and labour intensive. Especially the investigation in viticulture is restricted to an on-site analysis due to the perennial nature of grapevine. Traditionally skilled experts examine small samples and extrapolate the results to a whole plot. Thereby different grapevine varieties and training systems, e.g. vertical shoot positioning (VSP) and semi minimal pruned hedges (SMPH) pose different challenges.

In this paper we present an objective framework based on automatic image analysis which works on two different training systems. 
The images are collected semi automatic by a camera system which is installed in a modified grape harvester. The system produces overlapping images from the sides of the plants.
Our framework uses a convolutional neural network to detect single berries in images by performing a semantic segmentation. Each berry is then counted with a connected component algorithm.
We compare our results with the Mask-RCNN, a state-of-the-art network for instance segmentation and with a regression approach for counting.
The experiments presented in this paper show that we are able to detect green berries in images despite of different training systems.
We achieve an accuracy for the berry detection of 94.0\% in the VSP and 85.6\% in the SMPH.
\end{abstract}


\end{frontmatter}


\section{Introduction}
\label{S:1}
Grapevine is an important crop, historically and economically. In contrast to many other crops, the recent development in breeding is not faced towards yield enhancement, the goal is to breed robust varieties which satisfy the same quality standards as the traditional ones \cite{Toepfer11}. 
To improve wine quality it is not desirable to grow as many grapes as possible, because a higher yield leads to a decrease in quality. Therefore thinning procedures are applied to produce high quality grapes. 

Grapevine is a perennial crop which means that the monitoring of the vines needs to be carried out in the field. Traditional phenotyping is performed by skilled experts who apply labour intensive and subjective methods (OIV\cite{OIV01}, BBCH\cite{Lorenz95}). The methods range from visual screening to counting manually or weighting.
The experts only sample small regions of large grapevine plots and extrapolate the results for the whole field which leads to an error prone estimation.

\begin{figure}[h]
\centering\includegraphics[width=0.8\linewidth]{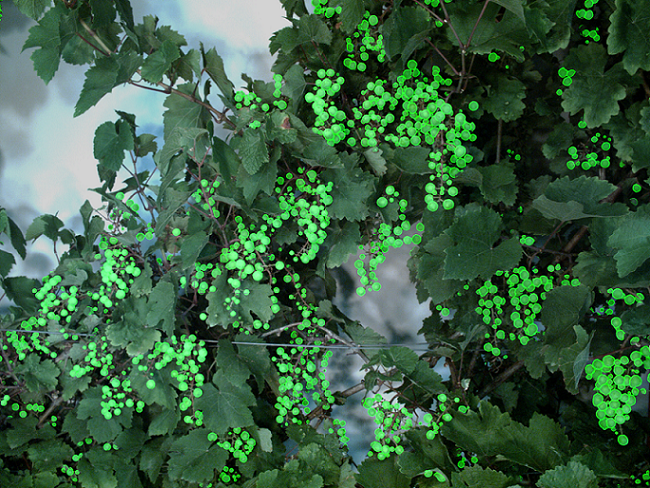}
\caption{Prediction after application of the neural network. The image shows a Riesling plant in the semi minimal pruned hedges training system. The resulting berry mask is put over the original image to give a visual impression of the result.}
\label{fig:Predi}
\end{figure}

The development of high-throughput phenotyping, the acquisition of large amounts of phenotypic data and the automatic analysis and extraction of phenotypic traits, was driven by the development of sensors and new image analysis techniques. RGB-, RGBD- or multi-spectral cameras can be used as well as laser scanners.
Gongal et al. \cite{Gongal15} did a review about different sensors and algorithms which were used to detect and localize fruits with a robotic background. 
Other authors use high-throughput phenotyping for crop breeding \cite{Araus14},  precision agriculture \cite{Kipp14}, the detection of diseases (e.g. \cite{Behmann15}, \cite{foerster2019}) or anomalies \cite{strothmann19} or the classification between different plant types to identify weeds (e.g., \cite{Milioto18_2}, \cite{lottes19}). Some of the main advantages of automatic procedures are objectivity, repeatability and high quality results.

Early approaches for high-throughput phenotyping for grapevine aimed at detecting grapes in images which were taken by handheld consumer cameras. The main focus lay in the recognition of geometrical structures.
This gives information about the spatial arrangement of the objects and in some cases even about the object size.
For example, the approaches used in \cite{Roscher13} and \cite{Nuske11} define berries as circular objects and use Hough transform or radial symmetry transform  to detect them. In \cite{Nyarko18}, convex surfaces are identified and used for fruit recognition.
 
Later Nuske et al. \cite{Nuske14} presented a large scale experiment for berry detection with a moving platform and semi automatic image acquisition with illumination in a realistic environment. They apply a circular Hough transform to detect berry candidates and classify them with texture, colour and shape features. Afterwards they cluster neighbouring berries into clusters.
Other approaches investigate the 3D structure of berries.
Rose et al. \cite{Rose16} used a stereo camera system to reconstruct point clouds from image sequences. They used colour and shape features to distinguish between canopy and berries.
\cite{Rist18} used a handheld laser scanner to produce high resolutions scans from bunches under laboratory conditions. They extract several parameters regarding single bunches.

Since 2012, with the work of Krizhevsky et al. \cite{Krizhevsky12}, neural networks (NN) became state-of-the-art for image classification tasks.
Moreover, by introducing convolutions to capture the spatial characteristics of objects in images and extending NN to convolutional neural networks (CNNs) \cite{Long15}, applications such as pixel-wise semantic segmentation were advancing. However, solving the task as semantic segmentation does not allow for a distinction of single instances in image regions, which is most obvious for neighbored image objects of the same semantic class. 
A distinction between single objects is realized extending a semantic segmentation task to instance segmentation, which outputs bounding boxes and several segmentation masks for single objects. One of the most famous approaches is called Mask-RCNN \cite{He17} where a major disadvantage is that the algorithms needs a predefined number of object proposals. 

In contrast to the detection approaches, several other ideas how to count objects in images exist.
One possibility is to count in an image without detecting their exact location. This is realized, for example, by regression methods as presented in \cite{Lemptisky10} and \cite{Cohen17}. 
The areas of application are diverse, from counting penguins in colonies \cite{Arteta16}, cells in microscopy images (e.g., \cite{Xie16}, \cite{Guo19}),  buildings in high-resolution over head images \cite{Lobry19} or nearly class agnostic approaches \cite{Lu18}. These works focus on avoiding explicit detection to count objects in images. Instead they output either a single number for each image or in some cases offer the possibility to retrieve spatial information while counting with the estimation of density maps.

In the field of remote sensing with images the automatic detection of buildings is often realized with the classical detection approach. 
Yang et al. \cite{Yang18} proposed the combination of SegNet with signed-distance labels to improve the detection of building from images. Marmanis et al. \cite{Marmanis18} on the other hand refine their buildings by adding information from an edge detector network.

An overview about the usage of neural networks and deep learning in agricultural applications was done by Kamilaris et al. \cite{Kamilaris2018}. Similar techniques were used by several researcher regarding the problem of yield estimation of grapevine. Aquino et al. first proposed a smartphone application where single bunches had to be surrounded by a black background \cite{Aquino16}.They detected circular light reflection and classified the results with a neural network. Later they disregarded the need for a background box \cite{Aquino17}. Another approach aimed at detecting regions containing grapevine inflorescences in images with neural networks \cite{Rudolph18} and applying a circular Hough transform in a second step. An adaptive network for semantic segmentation which was evaluated on different growth stages and data sets was proposed by \cite{Grimm19}. They either detect regions containing berries or dotwise berry positions.
An example for instance segmentation via Mask-RCNN for grapevine was done by \cite{Nellithimaru19}. They applied the Mask-RCNN to images which were simultaneously used for a 3D reconstruction.

We present a novel and objective approach to determine the number of berries as a decision base for thinning methods by providing berry numbers for whole rows.
The data collection is performed with a modified grapevine harvester called Phenoliner \cite{Kicherer17}. The harvesting equipment is replaced by a camera system which continuously records images laterally from the canopy while the harvester drives along the rows.

\begin{figure*}[h]
    \centering
    \begin{subfigure}[b]{0.48\textwidth}
        \centering
        \includegraphics[height = 7cm]{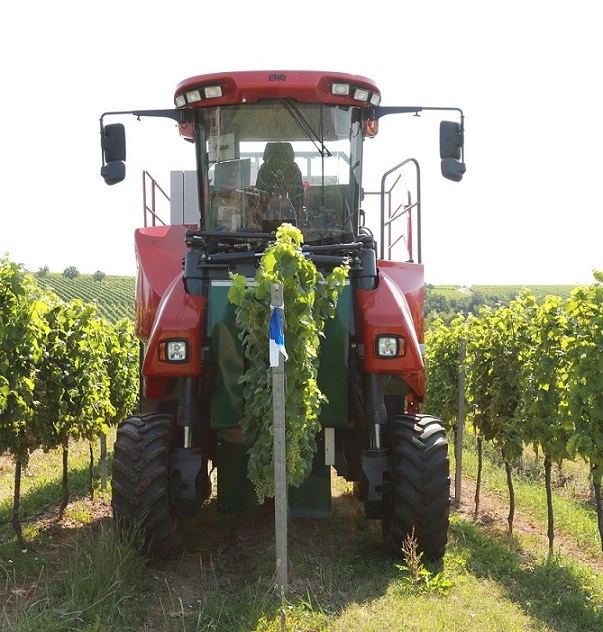}
        \caption[]%
        {{\small Phenoliner}}    
        \label{fig:Phenoliner}
    \end{subfigure}
    \quad
    \begin{subfigure}[b]{0.48\textwidth}  
        \centering 
        \includegraphics[height = 7cm]{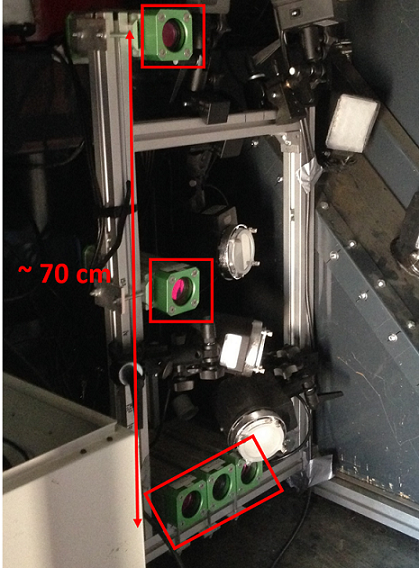}
        \caption[]%
        {{\small Camera System}}    
        \label{fig:CameraSyst}
    \end{subfigure}
    \caption[ The average and standard deviation of critical parameters ]
    {\small \ref{fig:Phenoliner} shows the phenotyping platform "Phenoliner", which was first introduced by Kicherer et al. \cite{Kicherer17}. It is based on a grapevine harvester where the harvesting equipment is replaced by a camera system. The system can be seen in \ref{fig:CameraSyst} and consist of 5 cameras which deliver overlapping images of the canopy. The vertical cameras are positioned with 35 cm between each camera resulting in a maximum distance between the outer two cameras of approximately 70 cm. 1.2 m of the canopy are covered vertically. 
    \label{fig:R2}}
\end{figure*}

To avoid a computationally intensive instance segmentation we reformulate a semantic segmentation task in a way that results in single object instances without object proposals. We define three classes, 'berry', 'edge' and 'background' so that every single berry is separated from neighbouring berries by an edge and can therefore be identified as a single instance. There is no need to explicitly perform an instance segmentation.\\

The contributions of this paper are the following:
\begin{itemize}
    \item we present a novel, accurate and efficient way of counting by reformulating an instance segmentation into a semantic segmentation by introducing an additional class 'edge'
    \item we evaluate our algorithm thoroughly for two different pruning systems and show that our algorithm handles both convincingly
    \item we compare our algorithm with two different methods. First a state-of-the-art instance segmentation with Mask-RCNN. Secondly regression of a density map with U-Net.
\end{itemize}

\section{Materials and Methods}
\subsection{Data}
Our data set is part of a big campaign which was carried out in 2018 in an experimental vineyard at the JKI Geilweilerhof located in Siebeldingen, Germany. Data were acquired on three different dates in 2018. The first images were taken before the application of a thinning procedure, the second set of images was taken shortly after the thinning. Further images were taken shortly before harvest.

We observed two different training systems, the vertical shoot positioned (VSP) and the semi minimal pruned hedges (SMPH). The two training systems feature diverse difficulties. 

\begin{figure*}[h]
    \centering
    \begin{subfigure}[b]{0.44\textwidth}
        \centering
        \includegraphics[height = 6cm]{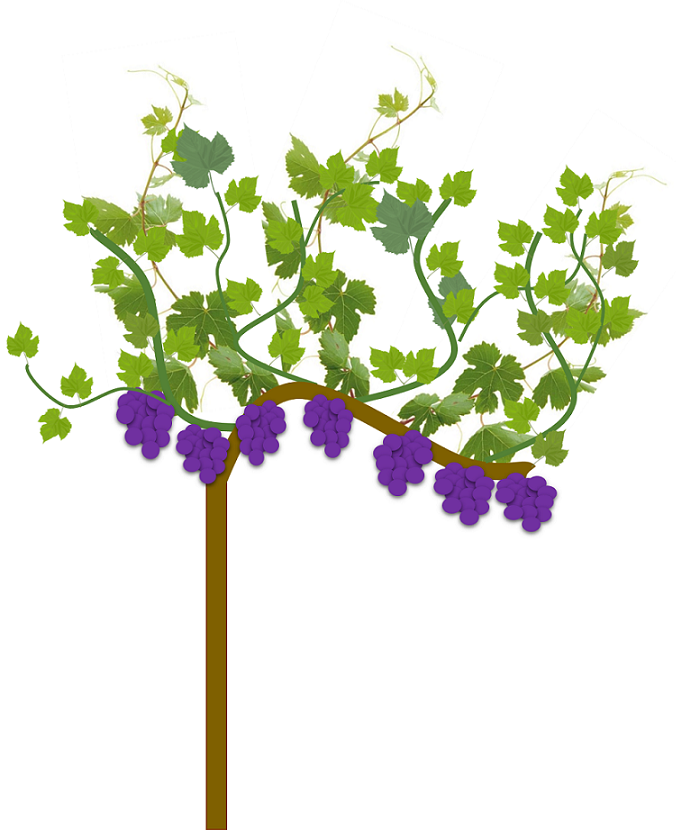}
        \caption[]%
        {{\small Vertical Shoot Positioned\newline (VSP)}}    
        \label{fig:VSP}
    \end{subfigure}
    \quad
    \begin{subfigure}[b]{0.52\textwidth}  
        \centering 
        \includegraphics[height = 6cm]{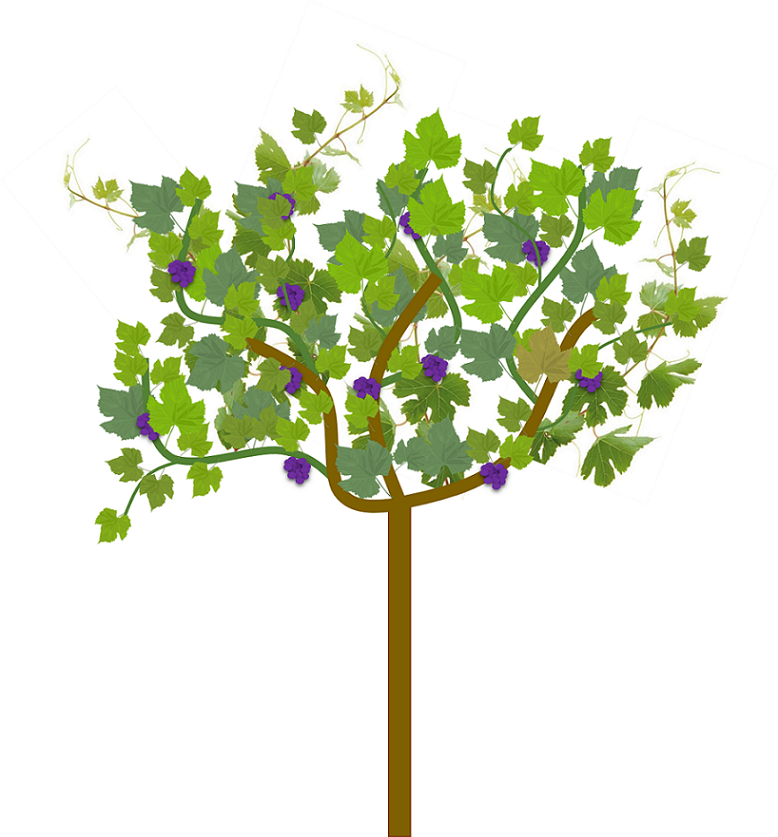}
        \caption[]%
        {{\small Semi Minimal Pruned Hedge\newline (SMPH)}}    
        \label{fig:SMPH}
    \end{subfigure}
    \caption[ The average and standard deviation of critical parameters ]
    {\small Depiction of the two different training systems. Fig. \ref{fig:VSP} shows an example for the vertical shoot positioned (VSP) system. It features one main branch which grows over multiple years while the others are removed annually. The grape bunches mainly grow in the bottom part of the canopy and feature a compact and homogeneous structure. 
    In contrast Fig. \ref{fig:SMPH} shows the semi minimal pruned hedges (SMPH). More branches are allowed to grow and the canopy is thicker. The grape bunches are positioned all over the canopy but often grow in the top part. The bunches itself are smaller, looser in structure and the berry size is inhomogeneous.
    \label{fig:TrainingSystems}}
\end{figure*}

The VSP is the traditional training system (see Fig. \ref{fig:TrainingSystems}). It features one main branch with several thinner shoots branching off (see Fig. \ref{fig:VSP}). Small branches and leafs are drastically reduced at the end of each season. Grape bunches mainly occur in the bottom part of the canopy, are seldom covered by leafs and feature a compact and homogeneous berry structure. 

The SMPH has a thick canopy which occludes many bunches (see Fig. \ref{fig:SMPH}) . Due to the minimal pruning more than one main branch exists. The grape bunches are spread through the whole canopy although they mainly occur in the upper part of the plant. The bunches itself have a loose structure and the berries have inhomogeneous sizes.

In each training system three different wine varieties were observed, namely Riesling, Felicia and Regent. The first two are white varieties while Regent is a red one. All varieties are part of the training set but for the evaluation we focus on the variety Riesling (see Fig. \ref{fig:Predi}).

\subsection{Sensor System}
We acquired images of grapevine with a field phenotyping platform called Phenoliner \cite{Kicherer17} which is shown in Fig. \ref{fig:Phenoliner}. The Phenoliner is a modified grapevine harvester. The harvesting equipment is removed and replaced by a camera and illumination system. 
The system consists of 5 cameras which can be seen in Fig. \ref{fig:CameraSyst}. 
3 RGB cameras are vertically aligned to cover the canopy of each vine. 
2 Additional cameras are installed in alignment with the bottom camera, building an L-Shape. 
A near-infrared camera is positioned in the middle of the horizontal cameras while the surrounding ones are RGB cameras.
The vertical cameras allow a 3D reconstruction, but this is not in the scope of this paper. 
Further equipment of the Phenoliner consists of a real-time-kinematic (RTK)-GPS system which enables the geo-referencing of each image. 
The cameras are triggered simultaneously since they are synchronized with the GPS clock.
The geo-reference allows the identification objects which occur more than once in overlapping images, though this is out of scope for this paper.

The cameras have a distance of approximately 75 cm to the canopy which results in a coverage of 1.2 m of the canopy in vertical direction. Each image has dimensions of 2592 $\times$ 2048 pixels and has a spatial resolution in the real world of 0.3 mm. 
For further information about the camera system we refer the reader to \cite{Kicherer17}.

We observed 10 plants in both training systems with the three vertical cameras, each plant was covered with 3 overlapping images. Each image is processed individually.
Images featuring the VSP show on average 329 and a maximum of 890 berries per image. 
The number of berries per image is higher for the SMPH with 556 berries on average and a maximum of more than 1100 per image. 

\subsection{Algorithms}
\begin{figure*}[h]
\centering\includegraphics[width=1\linewidth]{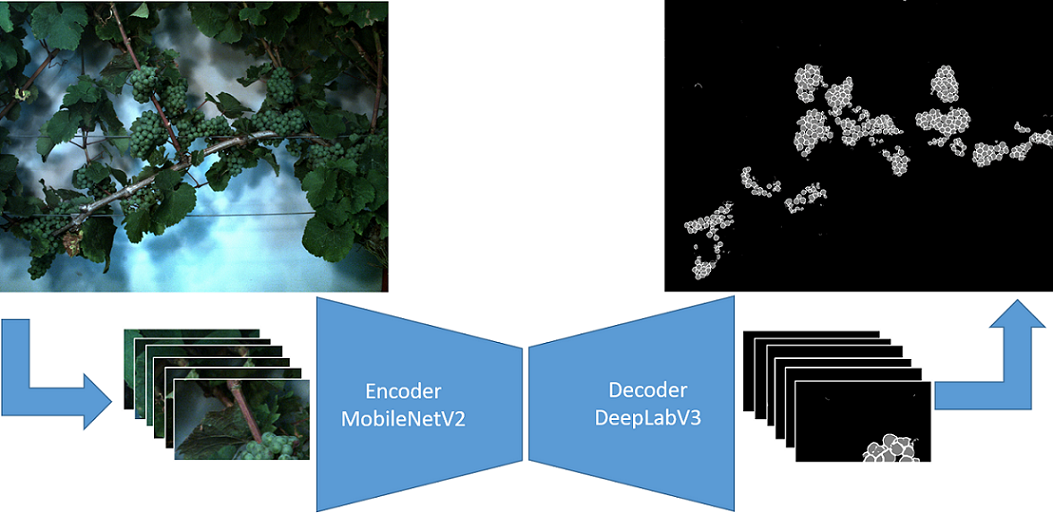}
\caption{Berry Segmentation Framework. For computational reasons each image is cut into overlapping image patches. Each image patch is classified by an encoder-decoder network. The encoder backbone consists of an MobilenetV2 \cite{Sandler18} and the decoder head of a DeepLabV3+ \cite{Chen18}. The patches are reconstructed into an image mask. Due to the overlap we avoid border fragmentation by applying a majority vote on regions containing more than one patch.}
\label{fig:Net}
\end{figure*}

The main contribution of this paper is the reformulation of an instance segmentation. More specifically we tackle a counting task with a semantic segmentation. We evade the detection and segmentation of every berry instance in an image by turning it into a pixel-wise classification with the classes 'berry', 'edge' and 'background'. Fig. \ref{fig:Net} shows an inferred mask on the right side. \\

The cameras mounted on the Phenoliner record images in DSLR-quality. They have a high resolution which makes them expensive to work with in terms of memory consumption. 
Two ways exist to handle high resolution images with convolutional neural networks (CNNs), namely down sampling or cutting the image into patches. Due to the fine structure and small size of the berries, down sampling leads to performance losses resulting in missing berries or wrong classifications. Therefore we crop each image into overlapping patches. The overlap is important to minimize edge artifacts. We use a majority vote over each overlapping image region. The result is a full segmentation mask with the same size and resolution as the original image without hurting the performance.
This means we have to run the CNN multiple times per image which calls for a lightweight yet powerful network architecture to efficiently process each patch.

\subsubsection{Network Structure}

We use a traditional U-shaped decoder-encoder architecture for pixel-wise semantic segmentation. The encoder backbone is a MobileNetV2 \cite{Sandler18} which introduced the inverted residual concept. The network has mobile applications in mind and poses an efficient and lightweight feature extraction that produces close to state-of-the-art results for tasks like classification, detection, and segmentation.
The decoder used is the DeepLabV3+ \cite{Chen18}. It refines the segmentation results with special focus on object boundaries.
The combination of encoder and decoder results in a fully convolutional semantic segmentation network. The framework is based on an open source implementation by Milioto et al. \cite{Milioto18}. 
We did not change the architecture of the model, but made adaptations to the data input, because we wanted to achieve the detection of single object instances with a semantic segmentation network. Our dataset definition is the focus of this work.
The network segments berries, edges and background accurately and performs fast on a moving platform. 
Keeping in mind that our application includes large amounts of data, we decide on a lightweight architecture which allows fast processing and decision processes.

\subsubsection{Loss Function}
As a loss function we use an Intersection over Union (IoU) loss as proposed by Yu et al. \cite{Yu16}. The IoU depicts the similarity between the prediction and the reference labels and can be defined as followed:

\begin{equation}
    \textrm{IoU} = \frac{|\textrm{A} \cap \textrm{B}|}{|\textrm{A} \cup \textrm{B}|} = \frac{\textrm{TP}}{\textrm{FP} + \textrm{TP} + \textrm{FN}}
\end{equation}

$\textrm{A} \cap \textrm{B}$ denotes the intersection and $\textrm{A} \cup \textrm{B}$ the union of two data sets, in our case the prediction and the reference masks. The second formulation is an intuitive representation with classical image analysis measures. TP means true positive, FP false positive and FN false negative. \\
We formulate the IoU as a loss as followed:

\begin{equation}
    L_{IoU} = -\ln \frac{|\textrm{A} \cap \textrm{B}|}{|\textrm{A} \cup \textrm{B}|}
\end{equation}
\subsubsection{Image Annotation}
\label{sec:Annot}
We have two different sets of annotated data. On the one hand we have a set of pixel wise annotated images which are used to train and evaluate the CNN. On the other hand we have a second set of dot annotated images to evaluate the counting of berries in a more extended fashion.\\

\textbf{Pixel wise Annotation}\\

\begin{figure*}[h]
    \centering
    \begin{subfigure}[b]{0.48\textwidth}
        \centering
        \includegraphics[width=\textwidth]{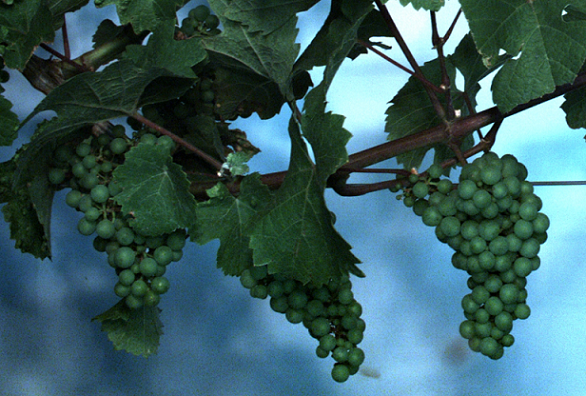}
        \caption[]%
        {{\small Original image}}    
        \label{fig:Orig}
    \end{subfigure}
    \quad
    \begin{subfigure}[b]{0.48\textwidth}  
        \centering 
        \includegraphics[width=\textwidth]{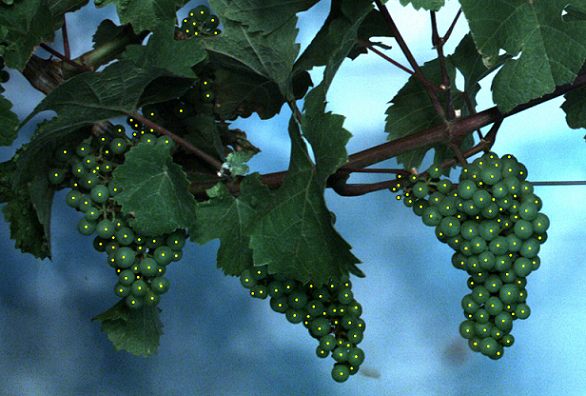}
        \caption[]%
        {{\small Dot wise annotation}}    
        \label{fig:DotAnn}
    \end{subfigure}
    \vskip\baselineskip
    \centering
    \begin{subfigure}[b]{0.48\textwidth}
        \centering
        \includegraphics[width=\textwidth]{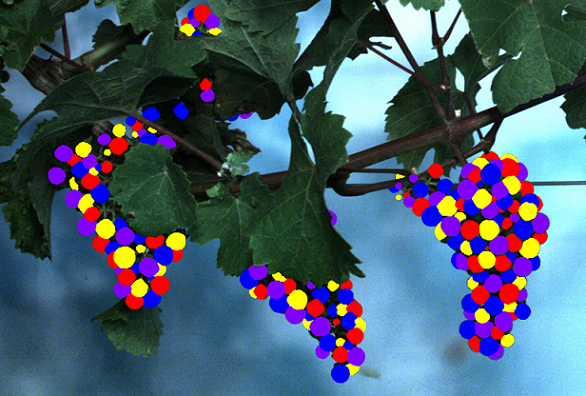}
        \caption[]%
        {{\small Original annotation}}    
        \label{fig:OrigAnn}
    \end{subfigure}
    \quad
    \begin{subfigure}[b]{0.48\textwidth}  
        \centering 
        \includegraphics[width=\textwidth]{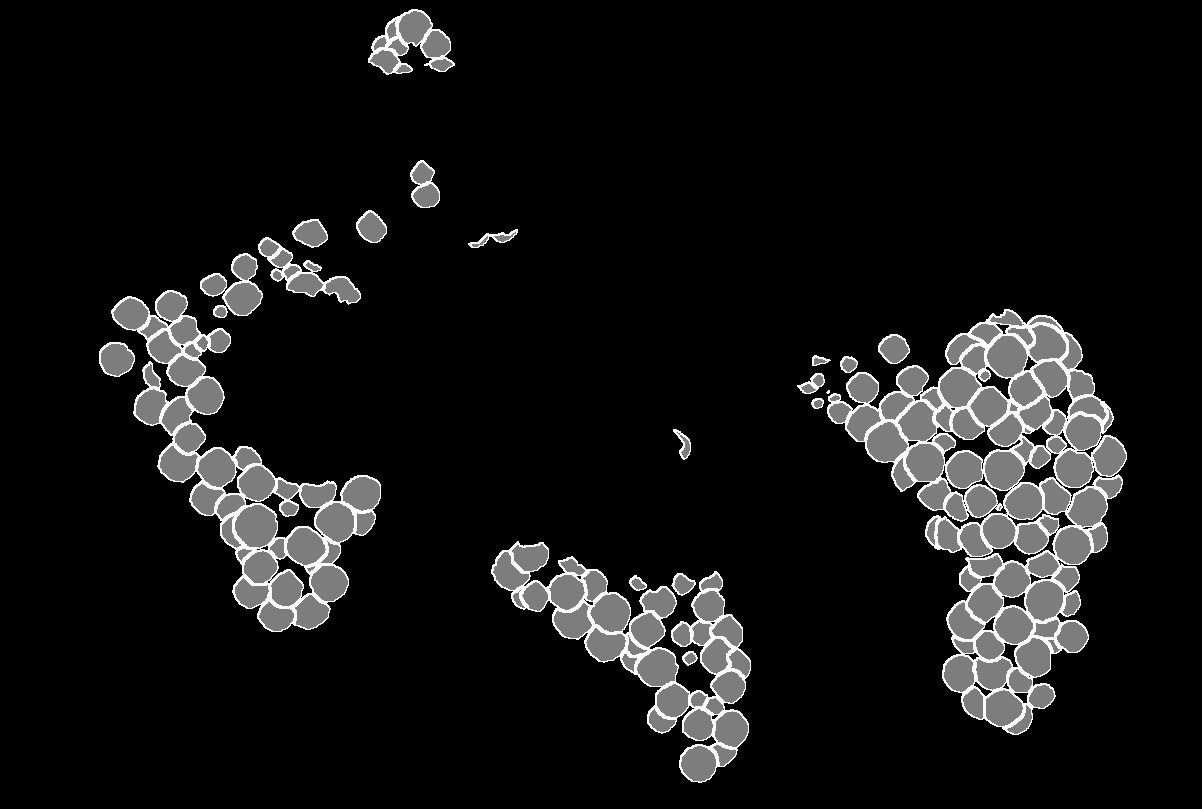}
        \caption[]%
        {{\small Berry-edge-format}}    
        \label{fig:BEAnn}
    \end{subfigure}
    \caption[ Annotation Process ]
    {\small Different stages of the annotation process. The first picture shows the original image without annotations. Fig. \ref{fig:DotAnn} shows an example of the dot annotated berries while Fig. \ref{fig:OrigAnn} demonstrates how the berries are originally marked with different colours. Later for each component an edge is computed (pixel width of edge is adjustable). 
    \label{fig:pixelAnnot}}
\end{figure*}

The detection of single berries in images is formulated as a semantic segmentation task with three classes: 'berry', edge' and 'background'. The labeling procedure consisted of colouring every berry in an image individually. Adjacent berries are labeled in different colours. From each berry component we compute the outer edge and label an edge with a fixed width (in our case 2 or 3 pixels). The edge width is a crucial parameter to distinguish between single elements. 
We use a fixed size due to simplicity reasons. In a single image scenario we don't have access to the depth information and can therefore not use an edge thickness which is depending on the distance to the camera. Furthermore the variation of berry sizes in different training systems has a higher impact than the size variation due to the distance from the camera.
The remaining inner parts of each berry are uniformly labeled into the class 'berry'. Everything else is denoted with the class 'background'. An example of the labeling and how the resulting mask with the berry edge formulation looks can be seen in Fig. \ref{fig:OrigAnn} and \ref{fig:BEAnn}.

We manually labeled 38 images showing different grapevine varieties and training systems to ensure a robust algorithm. 
The labeled data set features 61 \% images showing the SMPH and 30 \% show the VSP. This choice was done because the SMPH features more variety in the berry size and distribution. The occurrence of the different varieties is as followed: 55 \% Riesling, 23 \% Felicia and 23 \% Regent. 
The included images are from the two first recording times to show a variety of different grape sizes. The images show the plants before the veraison which means that all berries are green. \\

\textbf{Dot wise Annotation}\\

For evaluation purposes regarding the counting we created a set of dot wise annotated images. Each berry is manually marked with a dot. 
We annotated the images of 20 Riesling vines, 10 in the VSP and 10 in the SMPH. Each plant is covered by 3 overlapping images which leads to a total number of 60 dot annotated images. 
In Fig. \ref{fig:DotAnn} we can see an example for the dot annotations. In contrast to the pixel wise annotation for training this is a time efficient procedure and allows an extended evaluation of the counting.\\

\subsubsection{Post Processing}
To reduce the number of misclassifications we utilize prior knowledge about the geometry of berries. The main geometric property of berries is roundness although we investigate the quality of the prediction as well. We explore two possibilities to remove components which do not satisfy the definition of roundness. 

The initial steps of our post processing investigates the geometric properties via the minor and major axis of every component. This poses an intuitive definition of roundness. 
First we discard components which have a relation between minor axis $a_{min}$ and major axis $a_{maj}$ of less than $0.3$. This leads to a reduction of arbitrary shaped objects. This post processing stage is later called "Axis":

\begin{equation}
    \frac{a_{min}}{a_{maj}} > 0.3
\end{equation}

The second step focuses on the area of each component. We determine the radius of each component $\bar{r}$ by computing the mean between the minor and major axis. This radius is used to compute the theoretical area $A$ of a circle and compare it with the actual area $A_{comp}$ of the component. If the actual area $A_{comp}$ is more than 30 \% smaller than the computed circle area we discard the component. This leads to a reduction of leaf edges which are wrongly classified as berries, because of their crescent shape. This stage is later referred to as "Area":

\begin{equation}
    A < 0.3 \cdot A_{comp} = 0.3 \cdot \bar{r}^2 \cdot \pi = 0.3 \cdot \left(\frac{a_{min}+a_{maj}}{2}\right) ^2 \cdot \pi 
\end{equation}

A further visual investigation of the predictions shows, that correctly identified berries are often well surrounded by an edge. Misclassifications on the other hand are often insufficiently enclosed. Therefore we remove components which are surrounded by less than 40 \% and call the stage "Edge".

All parameters in the post processing step are chosen manually after performing several experiment with different values. We tried to achieve a suitable trade-off between removing misclassifications and not removing too many correct ones.
Nonetheless all parameters allow an intuitive understanding of the effects. 

\section{Results}
We thoroughly investigate our framework regarding different criteria and perform the following experiments:
\begin{itemize}
    \item analysis of intersection over union (IoU) as a classical error analysis for a semantic segmentation
    \item variation of edge thickness and its influence on the detection of berries
    \item analysis of post-processing steps and their influence on the detection of berries
    \item investigation of berry counting with $R^2$-plots
    \item comparison of the berry count with a classical instance segmentation approach, the Mask-RCNN
    \item comparison with U-Net which produces a density map.
    \item Qualitative evaluation of inference under different conditions
\end{itemize}

\subsection{Experimental Setup}
The network is trained on overlapping image patches. The patches are extracted from the 38 pixel wise annotated images (see first part of section \ref{sec:Annot}). Each patch has 432 $\times$ 256 pixels and a 50 \% overlap in vertical and horizontal dimension. 
We chose an overlap of 50 \% to cover each image region at least twice (at the edges of the image) and otherwise up to 4 times. This reduces the edge effects of the inferred masks. A higher overlap would be possible but results in a higher inference time due to the higher number of image pixels.
This results in a drastic reduction of training time in comparison to training on the full resolution of the images (2592 $\times$ 2048 pixels).

The data set contains 38 images, 90\% are used in the training set while 10\% are used for testing. This means that 4 images are chosen for testing, before we extract patches.
Furthermore we augment our whole data set to enhance the robustness of our network. We perform three different kinds of augmentations: flipping, blurring and gamma shifting. We flip the images only horizontally to preserve the characteristic that grape bunches get smaller in the lower part. The blurring is applied with random kernel sizes between 3 and 7 and the gamma shift is randomly chosen in between 0.8 and 1.2. 
Including data augmentations we end up with 5700 patches from 38 images.

We retrained a network which was pretrained on the imagenet dataset \cite{imagenet_cvpr09}. The learning rate of the network is 0.001 and the momentum is 0.9. The learning rate is decreased by 0.99 after 5 epochs.

\subsection{Intersection over Union (IoU)}
Furthermore we investigate the intersection over union (IoU) for every class. The IoU, also referred to as the Jaccard coefficient, is a similarity measure between two sample sets $A$ and $B$. The definition is the same as for our loss.
In our case one set is the reference mask and the other one is the prediction mask which is produced by the network. The evaluation is done for each class separately.
It is the most common measure to evaluate a semantic segmentation.

Tab. \ref{tab:Acc} shows that we achieve an IoU for the class 'berry' of more than 75~\% for both training systems. 
For the class 'edge' the IoU is more than 10~\% worse for 2 pixel edges compared to 3 pixels. The result is still reasonable because a thin edge is very hard to reproduce. 

\begin{table}[h]
\begin{center}
    \begin{tabular}{|l|c|c|c|c|c|}
        \hline
        & Edge [3 pix] & Edge [2 pix] \\
        \hline\hline
        Average IoU [\%]& 76.0 & 73.0 \\
        IoU Background [\%]& 99.0 & 99.1 \\
        IoU Berry [\%]& 75.3 & \textbf{76.8} \\
        IoU Egde [\%]& \textbf{53.7} & 42.0 \\
        \hline
    \end{tabular}
\end{center}
\caption{Investigation of the intersection over union (IoU). The IoU is better for berry than edge which can be explained with the nature of the classes. The overlay of thin edges is unlikely.}
\label{tab:Acc}
\end{table}

To further validate our model we also performed a cross validation on 30 images with an edge thickness of 2 pixels. For each batch we selected 3 images as validation images and the other 27 images as training images. We chose 30 images to evenly split our data. We chose the 30 images to be still representative for the whole network.
The average IoU is slightly worse than the above mentioned results with 69.96 $\pm$ 1.19~\%. The IoU for the background is 98.72 $\pm$ 0.45~\%, for the class 'berry' 72.03 $\pm$ 2.23 \% and for the class 'edge' 39.17 $\pm$ 2.41~\%. These values match the results from the fully trained network and indicate that our model fits well.

\subsection{Influence of Edge Thickness and Training System}
The class 'edge' has a major impact on the correct detection of single berries. The whole evaluation is done without the application of post processing steps.
We aim for an explicit differentiation between berries and examine two different edge thicknesses with this criterion in mind. We apply two different edge thicknesses, 2 and 3 pixels, on the two different training systems namely VSP and SMPH and evaluate the results in relation to the dot annotated images.
The precision $P$ describes the ratio between correctly predicted berries and all predicted berries. Correctly predicted means that a predicted berry region contains at least one manually annotated berry. The overall predicted berries contain the incorrectly detected berries as well, where no manual annotation lies within a component. 
\begin{equation}
    P = \frac{\textrm{TP}}{\textrm{TP} + \textrm{FP}}
\end{equation}
The recall $R$ describes the ratio between correctly predicted berries and all manually annotated berries. 
We have to keep in mind that one berry region can contain more than one manually annotated berry.
The $F_1$-Score is a measure for the test accuracy and contains the precision and the recall.

\begin{equation}
    R = \frac{\textrm{TP}}{\textrm{TP} + \textrm{FN}}
\end{equation}

\begin{table*}[h]
    \begin{center}
    \begin{tabular}{|l|c|c|c|c|}
        \hline
        \makecell{Training \\System} & Edge [pix] & Precision [\%] & Recall [\%]& $F_1$-Score\\
        \hline\hline
        VSP & 2 & \textbf{85.41} & \textbf{93.90} & \textbf{89.46}\\
        VSP & 3 & 81.21 & 92.59 & 86.53\\
        SMPH & 2 & \textbf{80.54} & \textbf{89.00} & \textbf{84.56}\\
        SMPH & 3 & 78.65 & 85.26 & 81.82\\
        \hline
    \end{tabular}
\end{center}
\caption{Comparison of various edge thickness values on different training systems. We show Precision, Recall and $F_1$-score.}
\label{tab:Edge}
\end{table*}
Tab. \ref{tab:Edge} shows that we recognize more berries in the VSP than in the SMPH. The improvement over different edge thicknesses can mainly be seen in the increased number of correct detections for the SMPH and in the decrease of the wrong classifications for both training systems.  This means that the influence of the edge thickness on the number of correct classifications is smaller for the VSP than for the SMPH.

The SMPH is characterized by an inhomogeneous berry size with a higher number of small berries compared to the VSP. Although we trained with an edge thicknesses of 2 or 3 pixels, the predictions feature thicker edges than the ground truth images. Due to the small berry size some of the smaller berries only consist of edge pixels if the thickness is too high. Our proposed method is then not able to recognize these small berries with a radius below 2 - 3 pixels.

\subsection{Influence of Post Processing}
\begin{figure*}[h]
    \centering
    \begin{subfigure}[b]{0.24\textwidth}
        \centering
        \includegraphics[width=\textwidth]{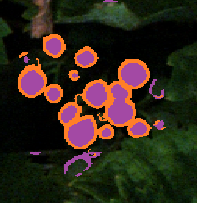}
        \caption[]%
        {{\small Original }}    
        \label{fig:OrigFilter}
    \end{subfigure}
    \begin{subfigure}[b]{0.24\textwidth}  
        \centering 
        \includegraphics[width=\textwidth]{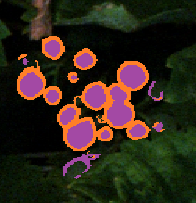}
        \caption[]%
        {{\small Axis}}    
        \label{fig:Axis}
    \end{subfigure}
    \begin{subfigure}[b]{0.24\textwidth}  
        \centering 
        \includegraphics[width=\textwidth]{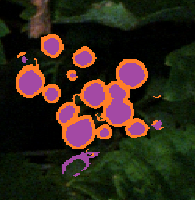}
        \caption[]%
        {{\small Area}}    
        \label{fig:Area}
    \end{subfigure}
    \begin{subfigure}[b]{0.24\textwidth}  
        \centering 
        \includegraphics[width=\textwidth]{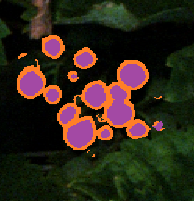}
        \caption[]%
        {{\small Edge}}    
        \label{fig:Edge}
    \end{subfigure}
    \caption[]
    {\small Results of the different filter stages. First we check the relation between major an minor axis. This removes objects which are not round. The second stage checks the relation between actual pixel area and the computed area of a circle with the diameter of the mean from the main axes. This stage removes objects which are round according to their main axes, but are not filled in. The last stage removes all components which are not sufficiently surrounded by an edge. We perform this computational intensive step as the last stage because we want to filter out as many objects as possible.
    \label{fig:Filter}}
\end{figure*}

The three post processing steps are applied to reduce the number of misclassifications, regions where a berry is detected, but no manually annotated berry is present. Fig. \ref{fig:Filter} shows an example of possible misclassifications and how the post processing steps reduces them.

Fig. \ref{fig:OrigFilter} shows the original prediction of the classes 'berry' and 'edge' overlayed with the original image. Artifacts at leaf edges can be seen as well as two berries which are fused into one component. 
The initial post processing step reduces components which have a major-minor-axis relation of less than 30 \%. The component which is removed is a small fragment in the left lower edge. 
On the right side a component remains which has a correct axis relation but is not filled in. In Fig. \ref{fig:Area} this component is removed because its actual area is smaller than the computed area of a circle with a radius computed from the mean of the main axis. 
The last and computational most intensive step is the removal of all components which are not sufficiently surrounded by an edge. Fig. \ref{fig:Edge} shows that the lowest component is removed due to this criteria. 

\begin{table*}[h]
    \begin{center}
        \begin{tabular}{|c|c|c|c|c|c|c|}
            \hline
            \makecell{Training \\System} & Axis & Area & Edge & Precision [\%] & Recall [\%] & $F_1$-Score\\
            \hline\hline
            VSP  & -   & -   & -   & 85.41 & \textbf{93.90} & 89.46\\
            VSP  & 0.3 & -   & -   & 88.32 & 93.67 & 90.92\\
            VSP  & 0.3 & 0.3 & -   & 88.92 & 93.65 & 91.22\\
            VSP  & 0.3 & 0.3 & 0.4 & \textbf{91.63} & 92.46 & \textbf{92.04}\\
            SMPH & -   & -   & -   & 80.54 & \textbf{89.00} &84.56 \\
            SMPH & 0.3 & -   & -   & 82.80 & 88.88 & 85.73 \\
            SMPH & 0.3 & 0.3 & -   & 83.49 & 88.81 & 86.07\\
            SMPH & 0.3 & 0.3 & 0.4 & \textbf{87.79} & 86.90 & \textbf{87.34}\\
            \hline
        \end{tabular}
    \end{center}
    \caption{Comparison of different filter strategies. Axis means that the relation between the minor and major axis of each component is not allowed to be smaller than 0.3. For the computation of a circle area we compute the radius of each component as the mean of the minor and major axis. We then compare the computed area with the actual area of each component. The actual area is not allowed to be smaller than 0.3 times the circle area. Edge means that every component needs to be surrounded by at least $40 \%$ of edge.}
    \label{tab:Filter}
\end{table*}

In table \ref{tab:Filter} we can see that for every filter stage the precision is increased while the recall decreases. Our filter removes a lot of misclassified berries but in some cases correctly classified berries are removed as well. 
Since the increase of the precision is stronger than the decrease of the recall, we remove more misclassified berries than correctly classified berries. 

\subsection{Evaluation of Berry Counting}
\begin{figure*}[h]
    \centering
    \begin{subfigure}[b]{0.48\textwidth}
        \centering
        \includegraphics[height = 6cm]{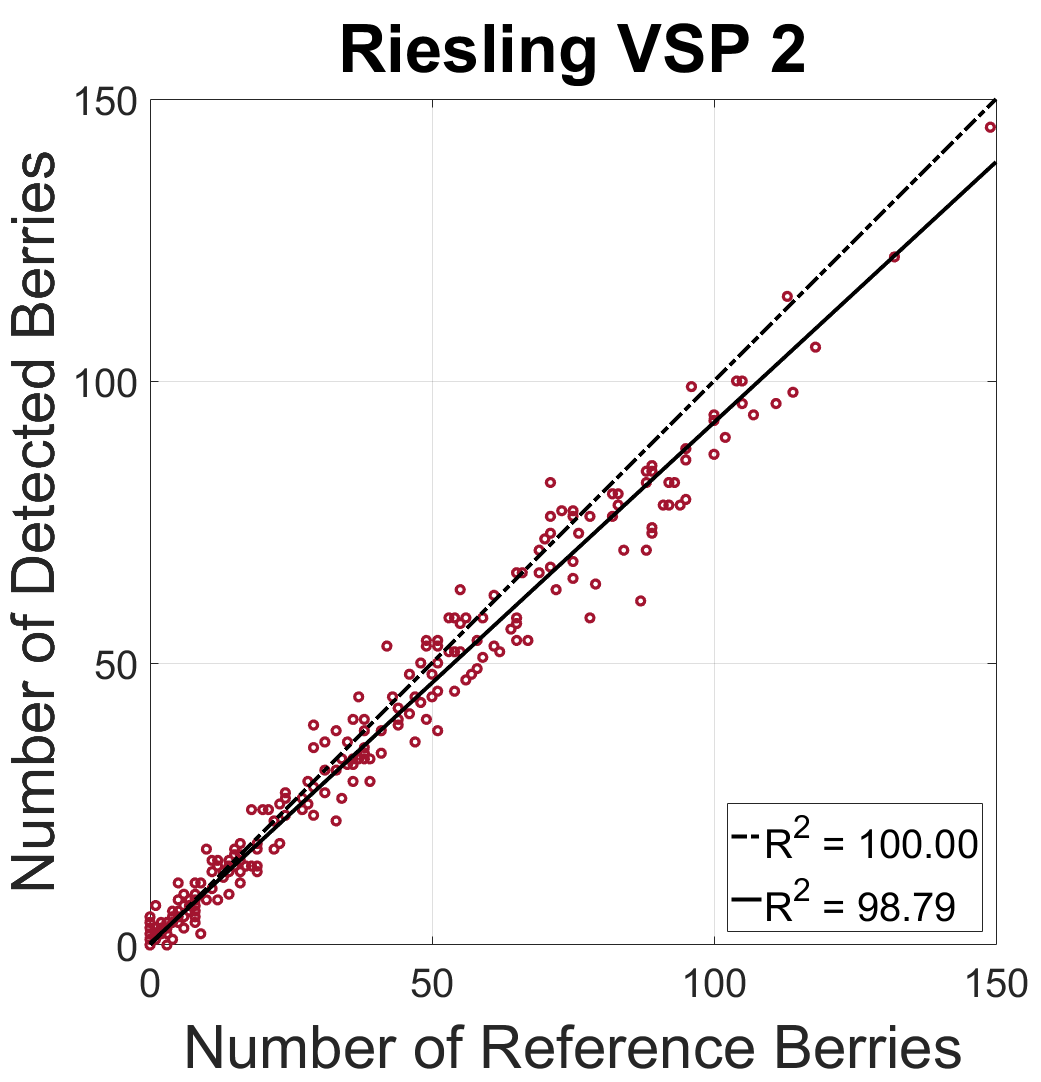}
        \caption[]%
        {{\small $R^2$-Plot for VSP ($R^2 = 98.79 \%$)}}    
        \label{fig:RVSP}
    \end{subfigure}
    \quad
    \begin{subfigure}[b]{0.48\textwidth}  
        \centering 
        \includegraphics[height = 6cm]{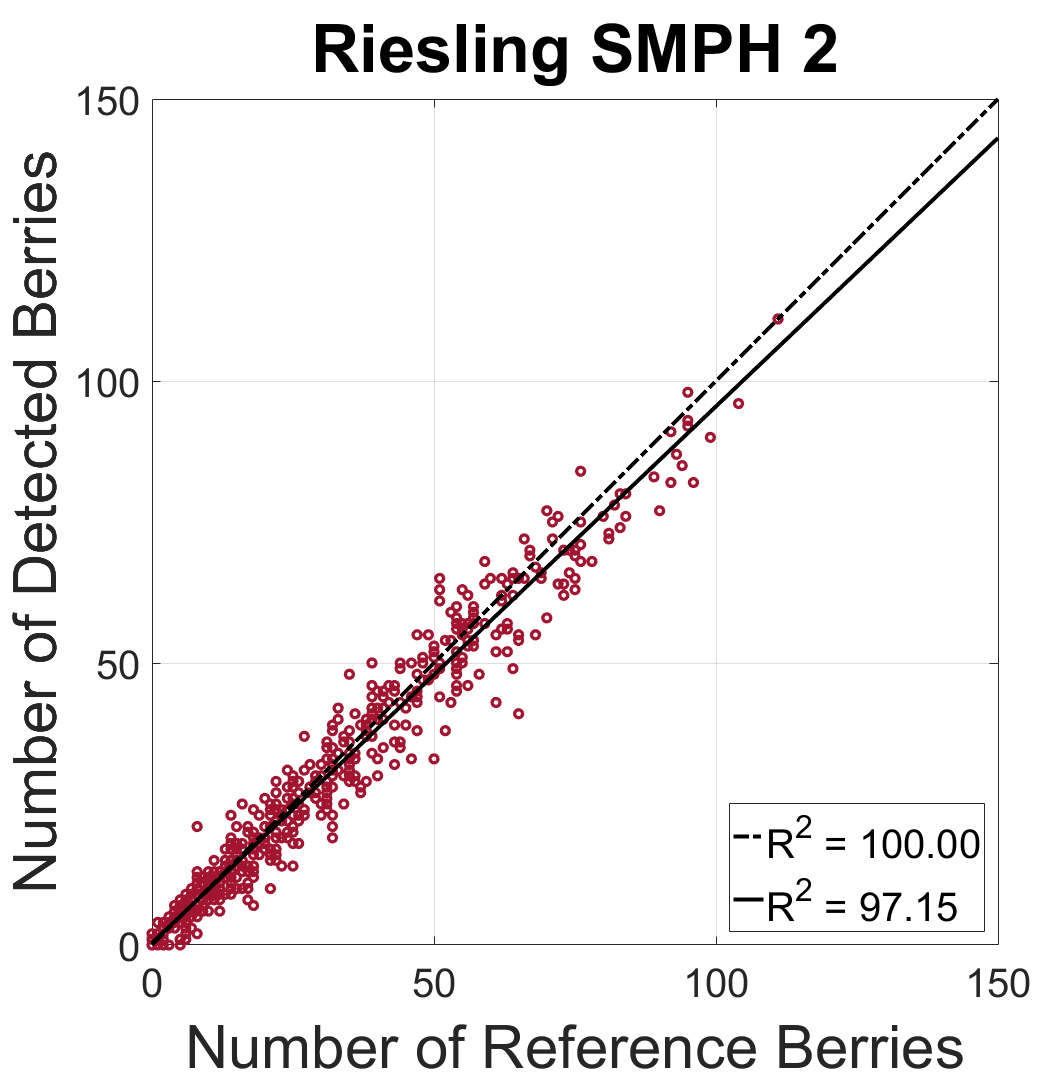}
        \caption[]%
        {{\small $R^2$-Plot for SMPH ($R^2 = 97.15 \%$)}}    
        \label{fig:RSMPH}
    \end{subfigure}
    \caption[ The average and standard deviation of critical parameters ]
    {\small $R^2$-Plots for the two different training systems VSP and SMPH. The red circles depict the berry count for an image patch. The x-axis shows the manual reference count, the y-axis the result of the connected component analysis of the predicted mask. The dashed line represents the optimal mapping between ground truth and prediction ($R^2 = 100 \%$). The continuous line is the estimated mapping, showing a good correlation in both cases.
    \label{fig:R2Plot}}
\end{figure*}

The proposed processing chain has the goal to count berries in the field. We therefore evaluate the complete chain by comparing its output, the predicted berry masks with the ground truth, our manually dot annotated berries. Especially in the agricultural science community this is often done by correlation plots.
The evaluation of the berry count is done on the dot annotated data set. It contains images of 10 plants in the VSP and 10 plants in the SMPH. Each plant is covered by 3 images which results in a total of 60 dot annotated images. 
We investigate the counting of berries by computing the coefficient of determination ($R^2$).
Fig. \ref{fig:R2Plot} shows the correlation plots for the VSP and the SMPH. The x-axis shows the number of manually counted berries while the y-axis shows the number of detected berries. Each point of the plots depicts the numbers for one non overlapping image patch cut from the images. The lines show the correlation between the number of reference and detected berries. The dashed line depicts a perfect correlation if we'd always detected the correct number of berries for each image. The continuous line is the actual correlation. 
We tend to underestimate the number of berries. The VSP shows with a $R^2$ of 98.79 \% a slightly better  correlation than the SMPH with $R^2$ of 97.15 \%. 

\subsection{Comparison with other approaches}

The task of object counting can be tackled with many different approaches. 
We want to show a comprehensive evaluation that compares our approach to both an instance segmentation and a regression approach where all approaches can be used to solve the same problem.
We compare our method against two well established methods.
The first one is an instance segmentation network called Mask-RCNN. The second one is a density map estimation with a U-Net.
Each network has a different structure, number of parameters and inference time (see Tab. \ref{tab:Networks}). Since our approach is very flexible in the sense that other segmentation networks can be used, for example U-Net, we still wanted to show that our method can be used successfully with a lightweight network architecture like the one proposed in Bonnet \cite{Milioto18}.

The two networks are trained and tested on the same data as our approach.

\begin{table*}[h]
    \begin{center}
        \begin{tabular}{|c|c|c|c|c|}
            \hline
            Network & Parameters Mio & Inference time& Spatial & Size \\
            \hline\hline
            Mask-RCNN & 64.159 & high& $\surd$ & $\surd$\\
            U-Net & 7.769 & low & $\surd$ & $\times$\\
            \hline
            Ours & 3.188 & low & $\surd$ & $\surd$\\
            \hline
        \end{tabular}
    \end{center}
    \caption{Comparison of different counting approaches. Mask-RCNN is a classical deep learning instance segmentation approach. It's a complex network with a high number of parameters and is able to detect object locations and the spatial extend of single objects. The U-Net is used to estimate density maps which give a count after integration. The spatial position of objects can be extracted but no information regarding their size. Our architecture with a MobileNetV2 encoder and a DeepLab3V+ decoder has the smallest number of parameter but is able to extract the spatial position and extend of single berries.}
    \label{tab:Networks}
\end{table*}

\subsubsection{Comparison with Mask-R CNN}
One of the most well known approaches for instance segmentation is the Mask-RCNN which was presented by He et al. \cite{He17}. Mask-RCNN is an extension of Faster-RCNN \cite{Ren15} which adds a third branch to the two already existing ones which provide a class label and a bounding box offset. The new branch outputs an object mask. The Mask-RCNN is therefore able to detect single objects in images by providing a mask for each one.

\begin{figure*}[h]
    \centering
    \begin{subfigure}[b]{0.35\textwidth}
        \centering
        \includegraphics[height = 3.3cm]{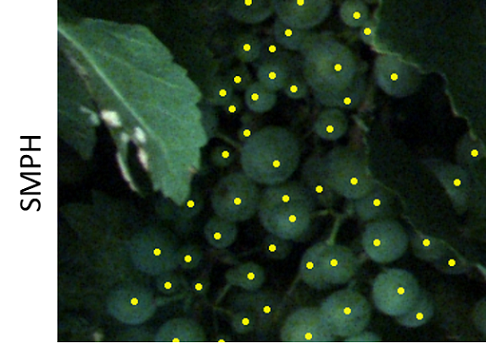}
        \caption[]%
        {{\small Manual Count}}    
        \label{fig:SMPH_Manu}
    \end{subfigure}
    \begin{subfigure}[b]{0.3\textwidth}  
        \centering 
        \includegraphics[height = 3.3cm]{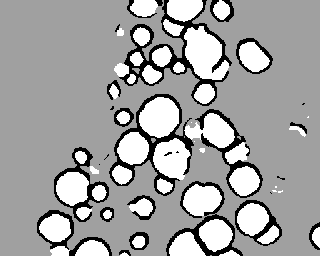}
        \caption[]%
        {{\small Own}}    
        \label{fig:SMPH_Own}
    \end{subfigure}
    \begin{subfigure}[b]{0.3\textwidth}  
        \centering 
        \includegraphics[height = 3.3cm]{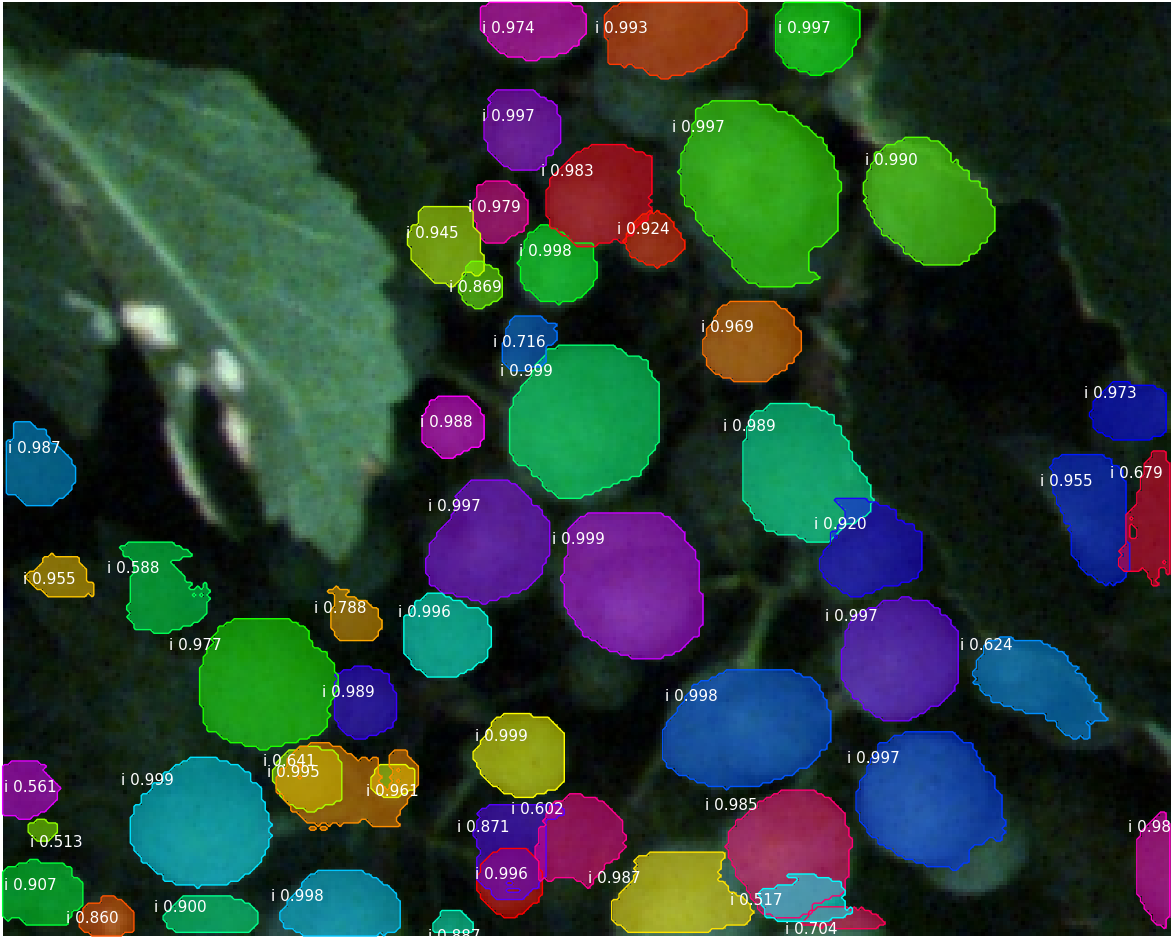}
        \caption[]%
        {{\small Mask-RCNN}}    
        \label{fig:SMPH_Mask}
    \end{subfigure}\\
    \begin{subfigure}[b]{0.35\textwidth}
        \centering
        \includegraphics[height = 3.3cm]{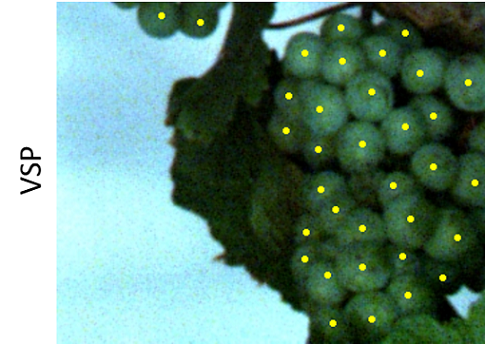}
        \caption[]%
        {{\small Manual Count}}    
        \label{fig:VSP_Manu}
    \end{subfigure}
    \begin{subfigure}[b]{0.3\textwidth}  
        \centering 
        \includegraphics[height = 3.3cm]{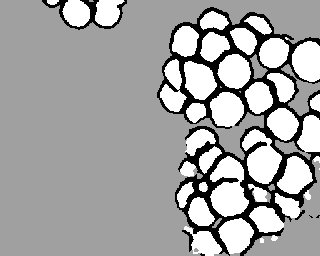}
        \caption[]%
        {{\small Own}}    
        \label{fig:VSP_Own}
    \end{subfigure}
    \begin{subfigure}[b]{0.3\textwidth}  
        \centering 
        \includegraphics[height = 3.3cm]{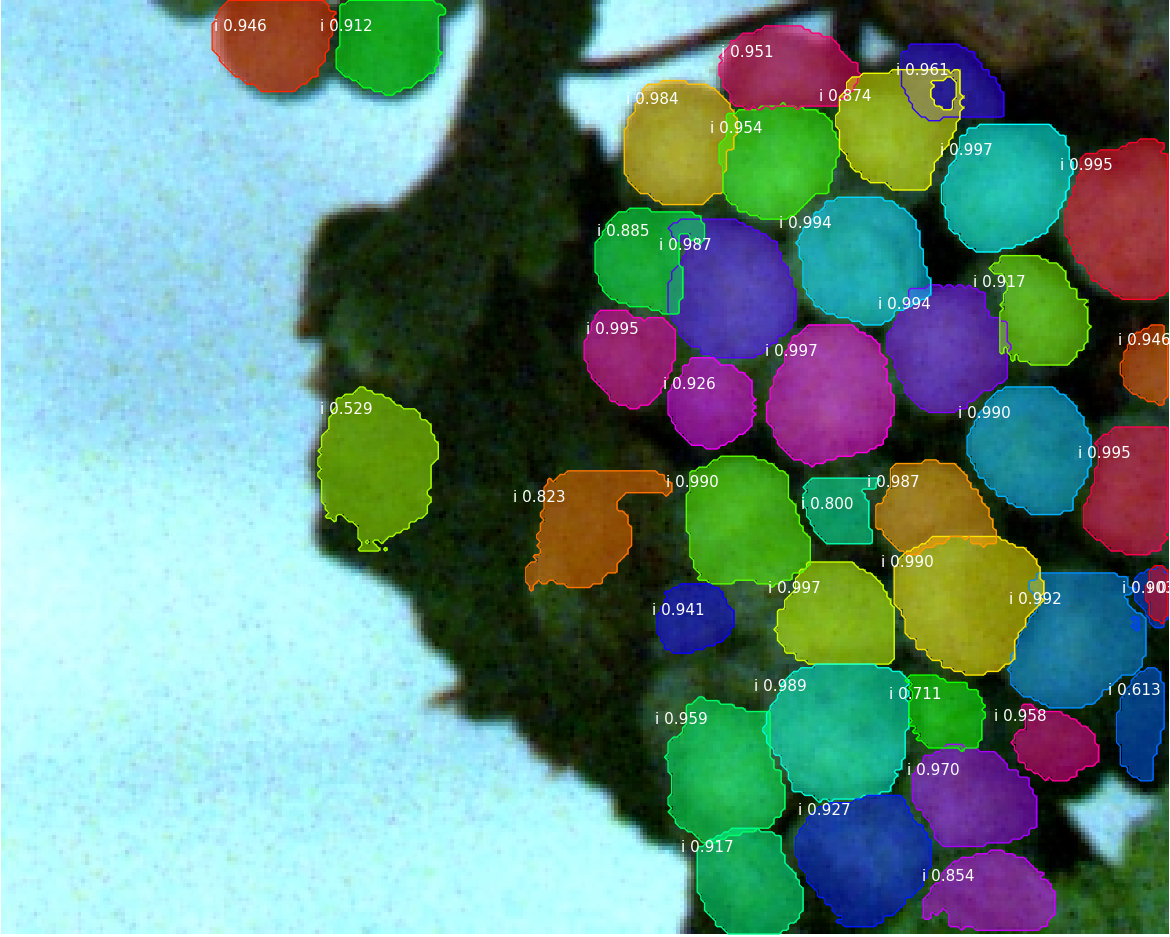}
        \caption[]%
        {{\small Mask-RCNN}}    
        \label{fig:VSP_Mask}
    \end{subfigure}
    \caption[ The average and standard deviation of critical parameters ]
    {\small Visual comparison of our own algorithm and the Mask-RCNN for an exemplary image patch for plant types SMPH (upper row) and VSP (lower row). In the first picture we can see the manual annotation for the counting reference. In the middle picture the mask which is outputed by our proposed algorithm is shown, while the last picture shows the result of the Mask-RCNN. The Mask-RCNN features more falsely detected berries in image regions where no berries should be detected.
    \label{fig:CompVisu}}
\end{figure*}

For training purposes we customize our already annotated data set in which every berry is individually coloured. In contrast to our 'edge' and 'berry' annotation we now consider each berry as a whole.
Each berry object presented by an own mask layer. In this layer only the classes 'berry' and 'background' exist. This means if we have 50 berries in an image we have 50 masks, one for each berry. These masks are stacked together to build a label matrix with the depth corresponding to the number of objects.

\begin{figure*}[h]
    \centering
    \begin{subfigure}[b]{0.35\textwidth}
        \centering
        \includegraphics[height = 4.3 cm]{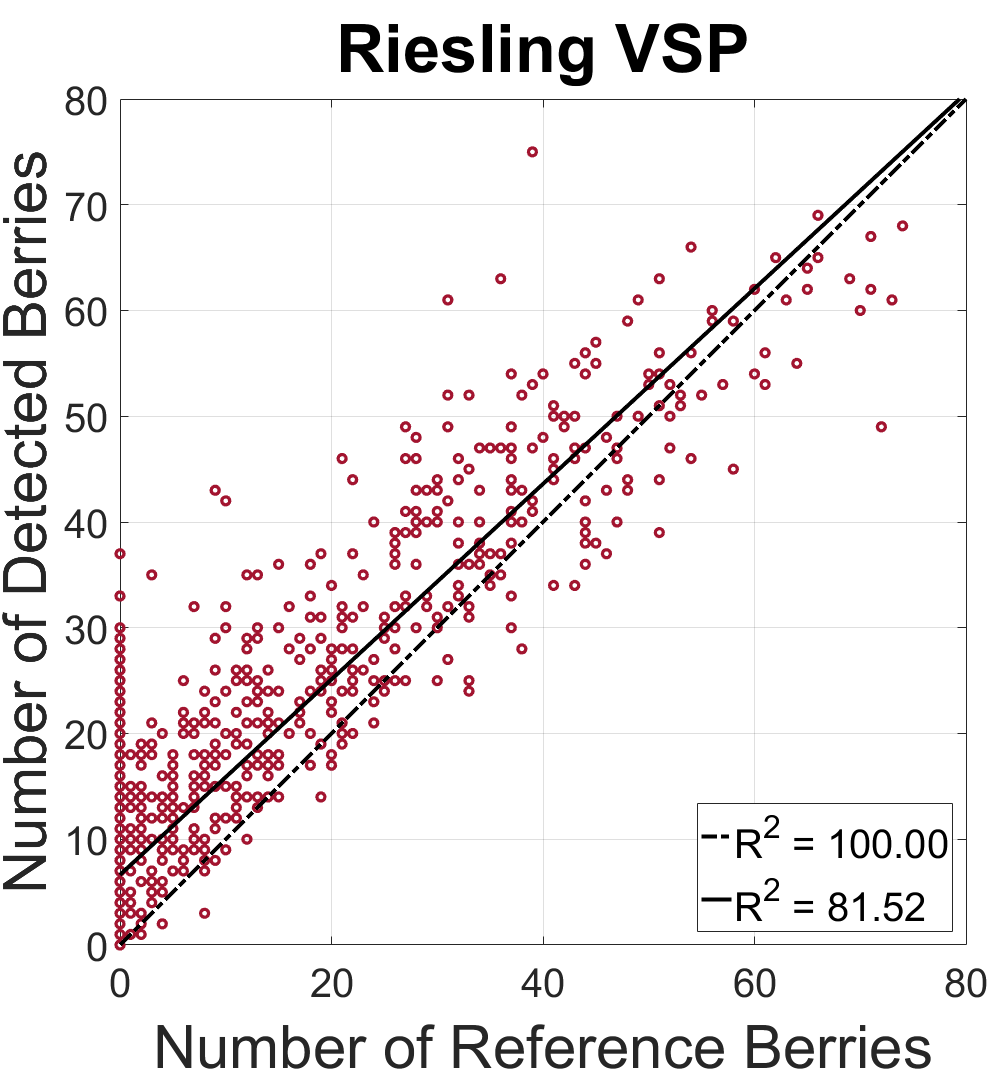}
        \caption[]%
        {{\small Mask-RCNN}}    
        \label{fig:R_Mask_VSP}
    \end{subfigure}
    \begin{subfigure}[b]{0.3\textwidth}  
        \centering 
        \includegraphics[height = 4.3 cm]{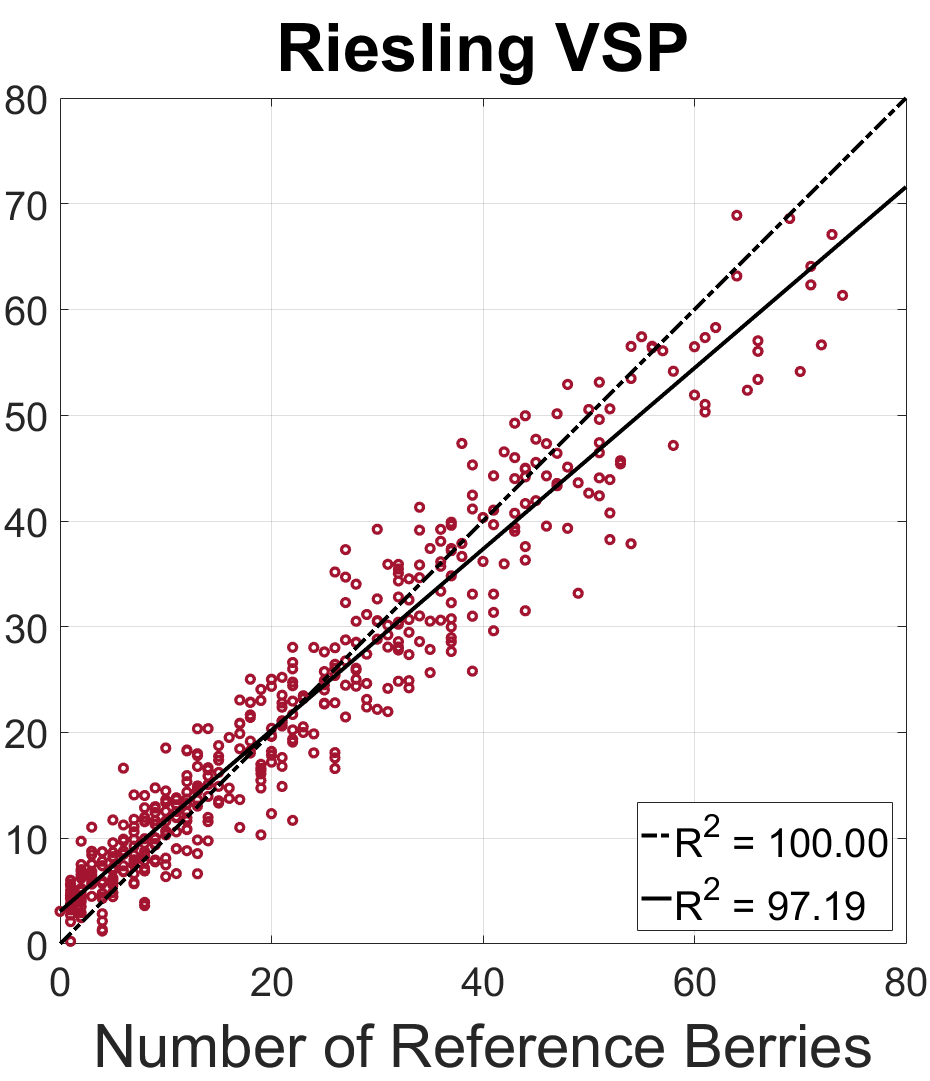}
        \caption[]%
        {{\small Regression}}    
        \label{fig:R_Reg_VSP}
    \end{subfigure}
    \begin{subfigure}[b]{0.3\textwidth}  
        \centering 
        \includegraphics[height = 4.3 cm]{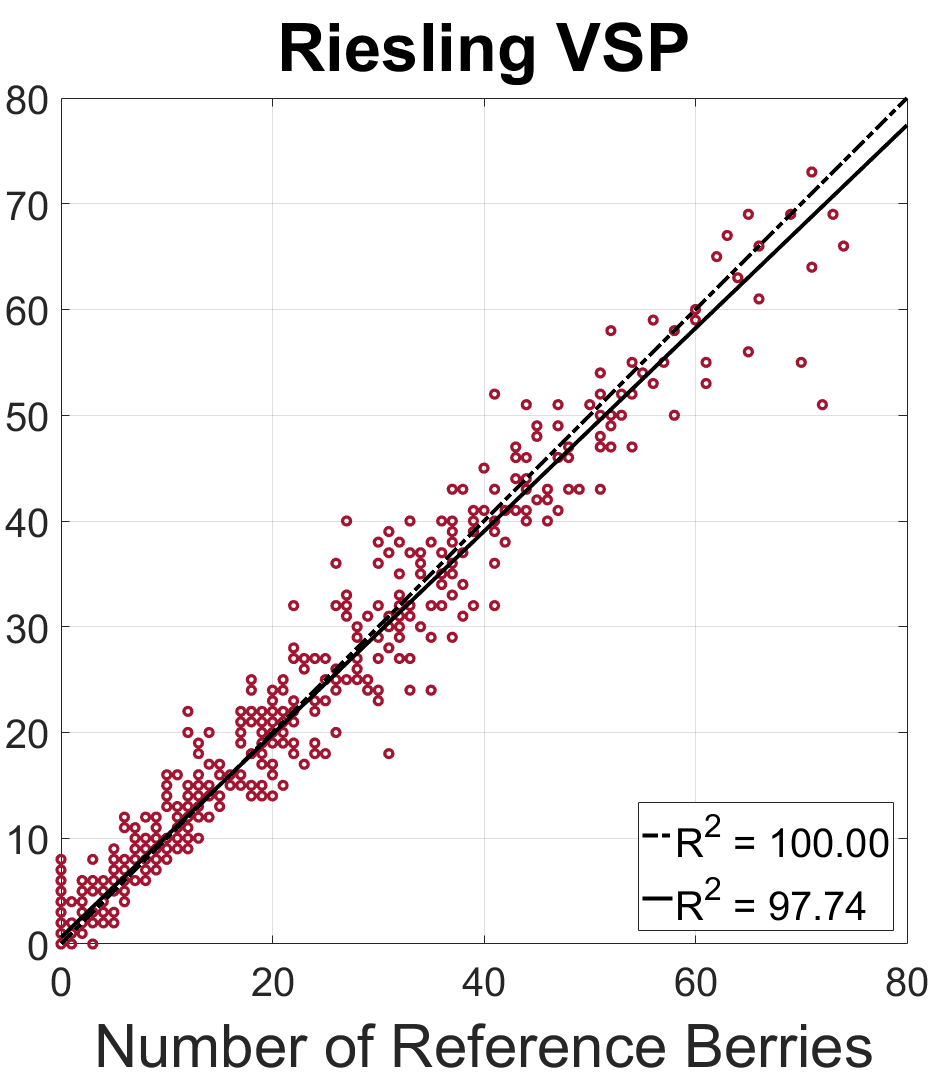}
        \caption[]%
        {{\small Own}}    
        \label{fig:R_Own_VSP}
    \end{subfigure}\\
    \begin{subfigure}[b]{0.35\textwidth}
        \centering
        \includegraphics[height = 4.3cm]{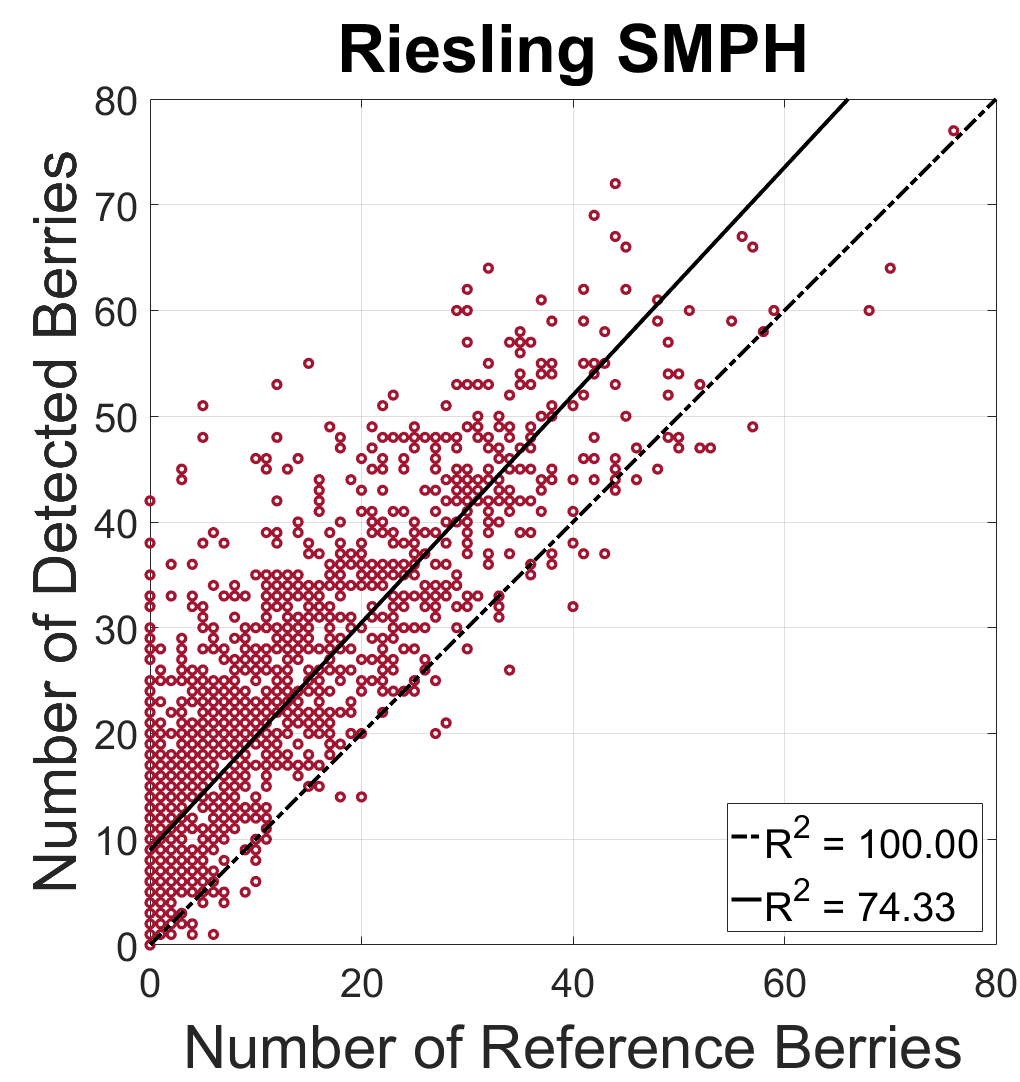}
        \caption[]%
        {{\small Mask-RCNN}}    
        \label{fig:R_Mask_SMPH}
    \end{subfigure}
    \begin{subfigure}[b]{0.3\textwidth}  
        \centering 
        \includegraphics[height = 4.3 cm]{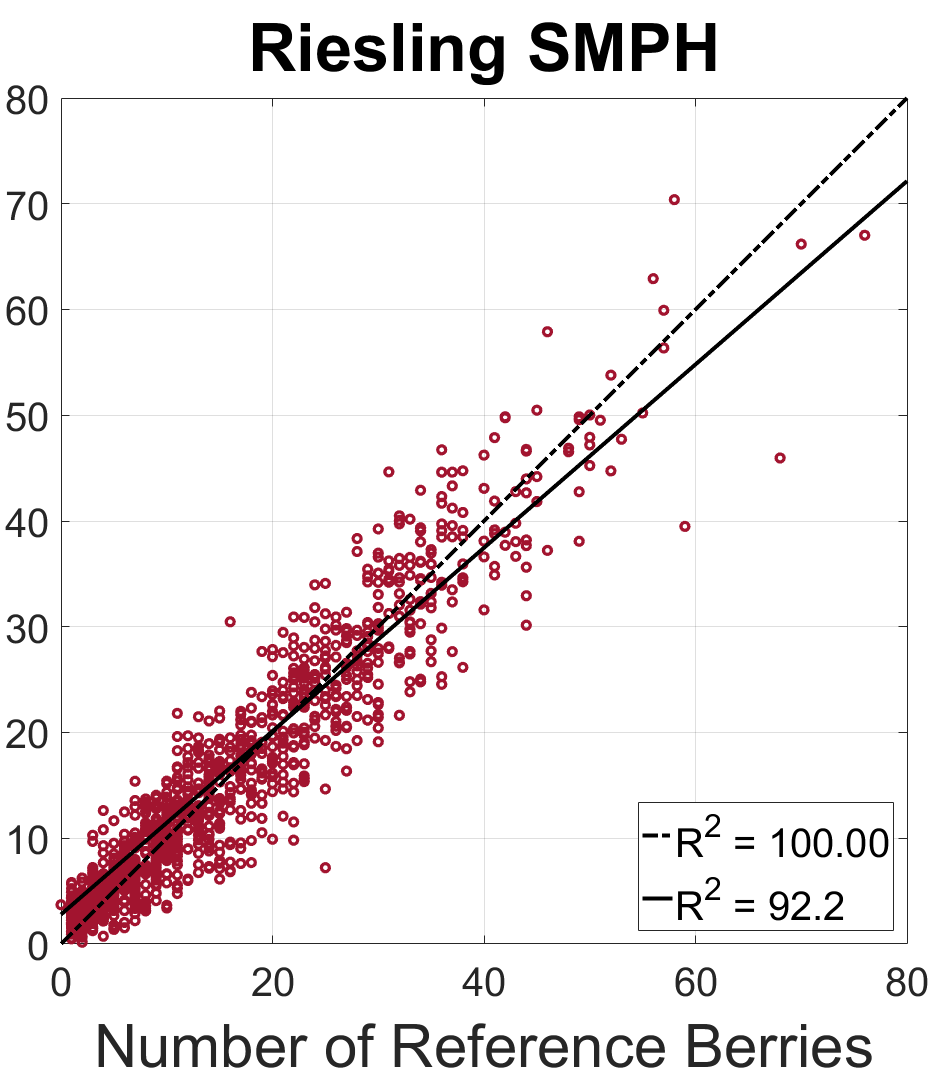}
        \caption[]%
        {{\small Regression}}    
        \label{fig:R_Reg_SMPH}
    \end{subfigure}
    \begin{subfigure}[b]{0.3\textwidth}  
        \centering 
        \includegraphics[height = 4.3 cm]{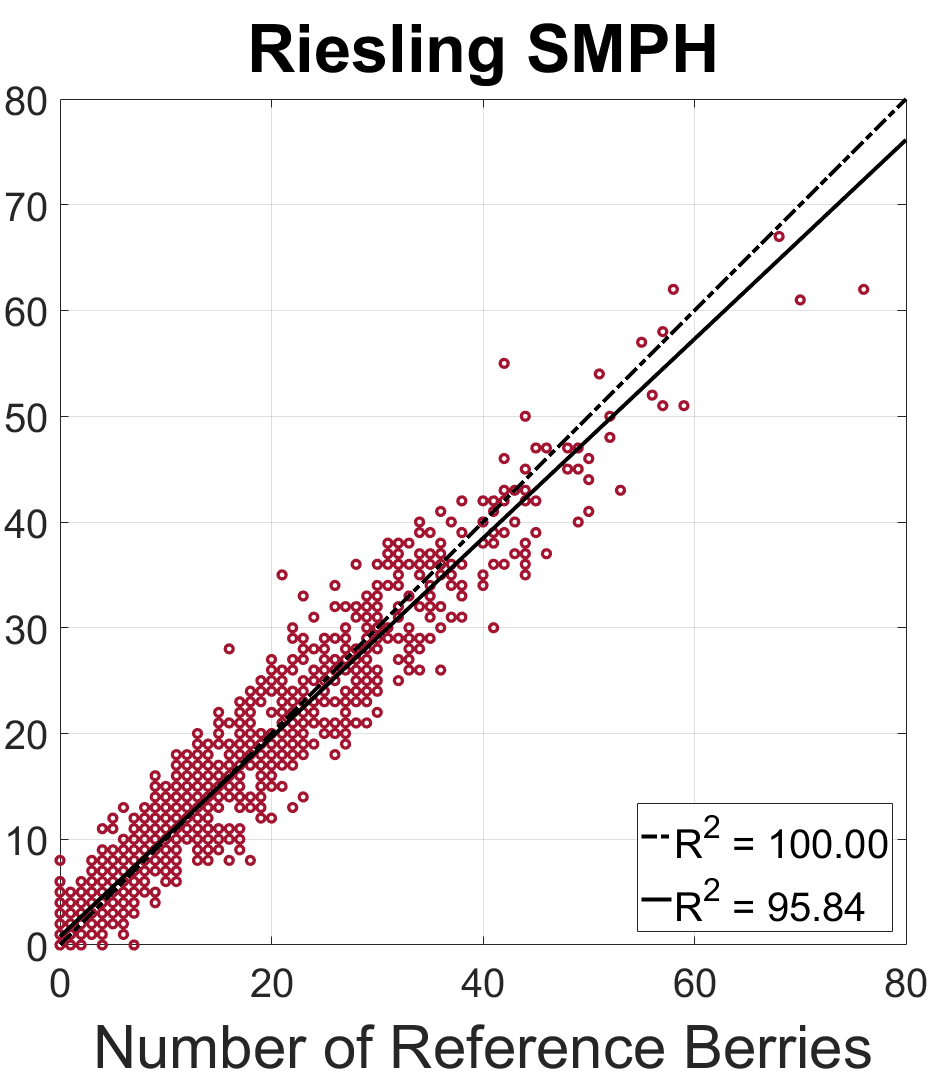}
        \caption[]%
        {{\small Own}}    
        \label{fig:R_Own_SMPH}
    \end{subfigure}
    \caption[ The average and standard deviation of critical parameters ]
    {\small Comparison of the berry count between Mask-RCNN, regression with U-Net and our approach. The Mask-RCNN overestimates the number of berries drastically for the SMPH as can be seen in Fig. \ref{fig:R_Mask_SMPH}. The $R^2$ is with 74.33 \% significantly worse than for our algorithms. The same is applicable for the VSP. The regression approach is only slightly worse than our approach.
    \label{fig:CompR}}
\end{figure*}

We use the Mask-RCNN implementation which is provided by Abdulla et al. \cite{Abdulla17}.
We train the network on image patches with the dimensions 320 $\times$ 256 pixels. The maximum number of objects is 80. Due to the small object sizes we use anchors of the sizes 8, 16, 32, 64 and 128 pixels with anchor ratios of 0.5, 1 and 2. 

To ensure a fair comparison between our algorithm and the Mask-RCNN we don't apply our post processing steps to the network outputs. Furthermore we train the network with the same patch size and number of images as the Mask-RCNN.
Fig. \ref{fig:CompVisu} shows a visual comparison between our algorithm and the results of the Mask-RCNN. For both training systems, the Mask-RCNN tends to overestimate the number of objects. 

We further investigate and compare the counting of berries in images with the coefficient of determination $R^2$ for our algorithm and the Mask-RCNN for both training systems. Fig. \ref{fig:R_Mask_SMPH} and Fig. \ref{fig:R_Mask_VSP} show the $R^2$-plots for the Mask-RCNN. In both cases the data points are very scattered. For our algorithm the data points are closer together and are approximated well by a straight line. The Mask-RCNN achieves $R^2$ between $74$ and $81~\%$ while our implementation shows a better correlation between the manually counted berries and the detected ones with $R^2 > 95~\%$.

There are several reasons which could be responsible for the bad performance of the Mask-RCNN. We only have a limited data set. We train our network on 5700 patches with a maximum number of 80 objects per patch or less. Since the Mask-RCNN has to train more than 64 Mio. parameters this might be not sufficient.

Another important aspect is the inference time. While we can infer our network on nearly 2000 images in roughly a minute, it takes the Mask-RCNN nearly an hour to process the same images. 
This differs by a factor of 60. Although we are mainly interested in the absolute inference time we have to keep in mind that the network architectures are highly disparate. The number of network parameters differs by a factor of 20, the number of parameters for the Mask-RCNN is approx. 64 Mio. while our network only has 3 Mio. The inference time decreases more than the parameter number increases.

\subsubsection{Comparison with U-Net density map estimation}
Besides the detection of objects, the problem of counting can be addressed by regression approaches as well. The idea is to  estimate a density map from an input image and to integrate over the map to retrieve the maxima in the density map. The goal of the procedure is to provide a count of the objects present in the image. We use a U-Net \cite{Ronneberger15}, an image segmentation network which is often applied for biomedical data.

The annotation process for regression networks is simpler than for detection network. Instead of pixel wise masks or bounding boxes a dot wise annotation is sufficient. We modify our data set of pixel wise annotated berries by providing the centers of each component as the dot annotation for each object. Around each dot a Gaussian Kernel with a deviation of one in both directions is applied.
We train the network\footnote{https://github.com/NeuroSYS-pl/objects\_counting\_dmap} on the same image patches as the Mask-RCNN and our own network. The patch dimensions are again $320 \times 256$ with a maximum number of 80 objects in each image.

The regression with the U-Net performs slightly worse than our approach. The $R^2$ for the VSP is $97.19 \%$ compared to $R^2 = 97.74\%$ from our approach. For the SMPH the difference is slightly larger with $R^2 = 92.20\%$ from the U-Net and $R^2 = 95.84\%$ from our approach.
The inference time is similar to our approach and U-Net has only two times as many parameters as our network (7.7 Mio).

The main advantage of our masks is the extraction of further phenotypic traits like the berry size. The regression approach yields density maps which could be used to extract the spatial positions of the single objects but not their whole extent.

\section{Qualitative investigation for difficult conditions}
\begin{figure*}[h]
    \centering
    \centering
    \begin{subfigure}[b]{0.48\textwidth}
        \centering
        \includegraphics[width=\textwidth]{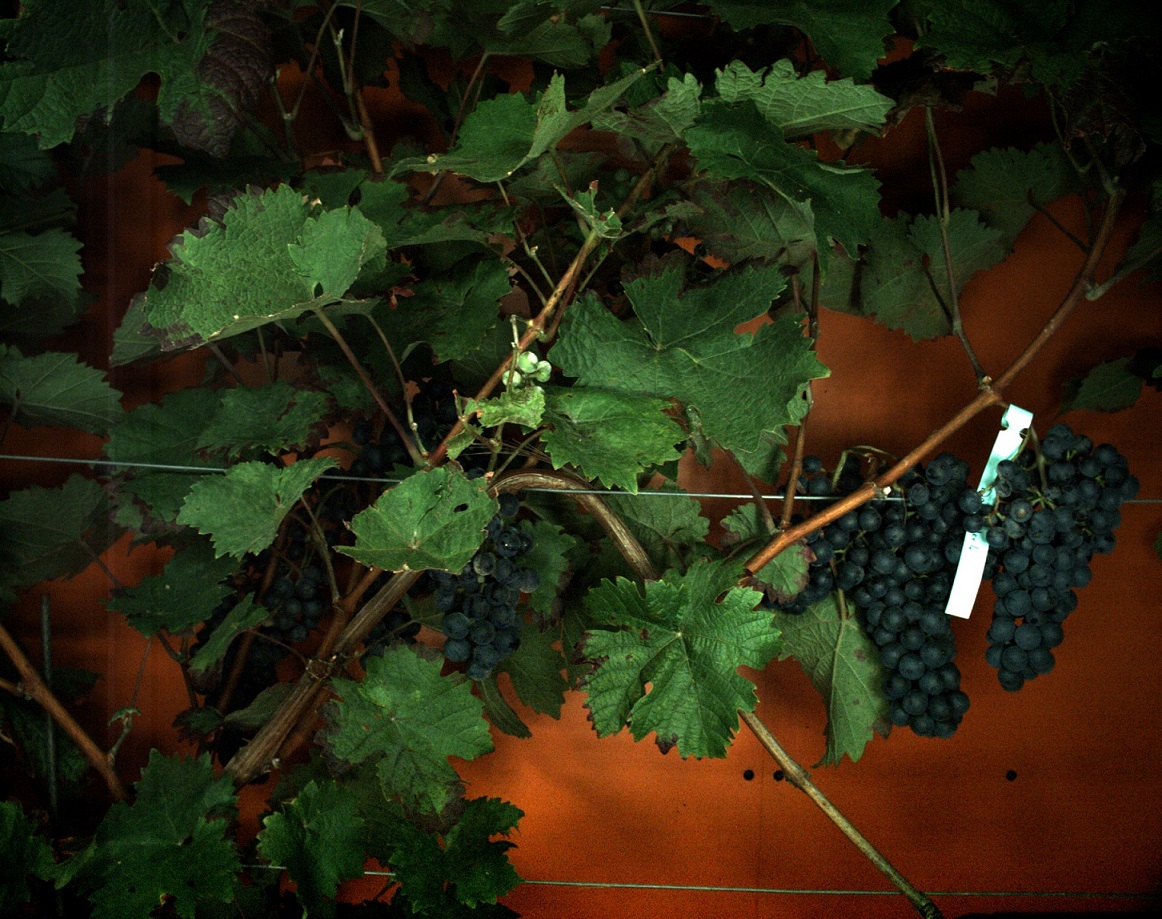}
        \caption[]%
        {{\small Phenoliner Image}}    
        \label{fig:PhenoOrig}
    \end{subfigure}
    \quad
    \begin{subfigure}[b]{0.48\textwidth}  
        \centering 
        \includegraphics[width=\textwidth]{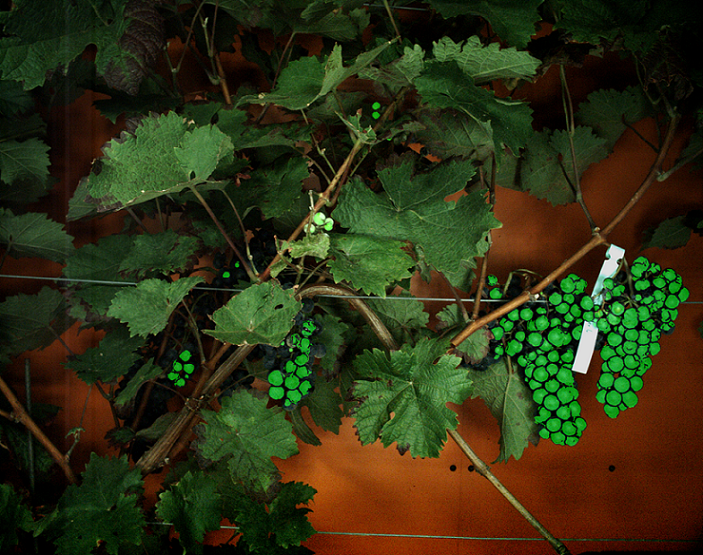}
        \caption[]%
        {{\small Prediction}}    
        \label{fig:PhenoPredi}
    \end{subfigure}
    \vskip\baselineskip
    \begin{subfigure}[b]{0.48\textwidth}
        \centering
        \includegraphics[width=\textwidth]{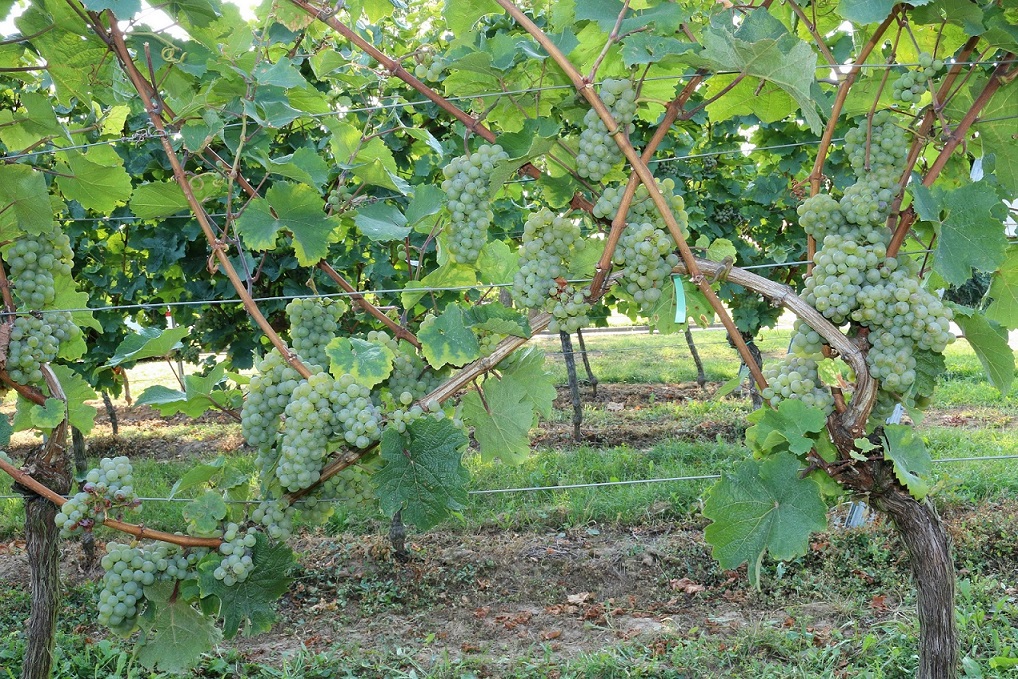}
        \caption[]%
        {{\small SLR Image}}    
        \label{fig:SLROrig}
    \end{subfigure}
    \quad
    \begin{subfigure}[b]{0.48\textwidth}  
        \centering 
        \includegraphics[width=\textwidth]{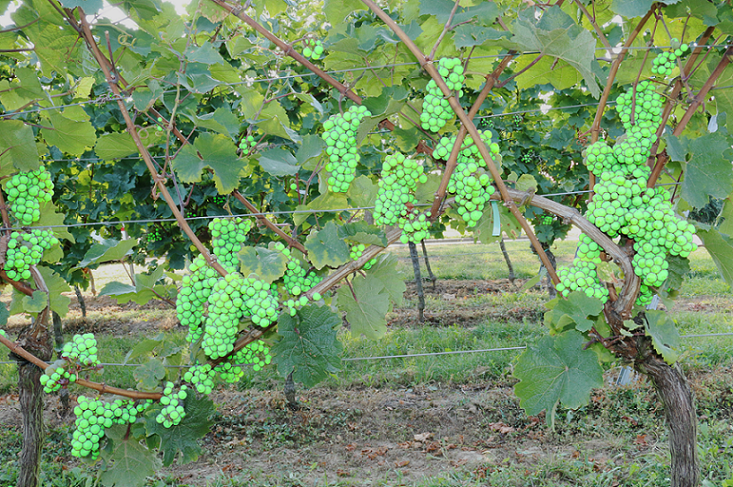}
        \caption[]%
        {{\small Prediction}}    
        \label{fig:SLRPredi}
    \end{subfigure}
    \vskip\baselineskip
    \caption[Outlook ]
    {\small Inference of our network on unrelated data. 
    Fig. \ref{fig:PhenoOrig} and \ref{fig:PhenoPredi} show images which were taken in 2017 with the Phenoliner. The illumination system and the background are different from the current setup. Furthermore the images show Regent, a red variety. Although the network never saw red wine berries it is able to detect them correctly.
    Fig. \ref{fig:SLROrig} and Fig. \ref{fig:SLRPredi} show a picture which was taken with a handheld camera under natural illumination. The network is nonetheless able to identify most of the berries.  
    \label{fig:Outlook}}
\end{figure*}

As an outlook, we show the ability of our network to adapt to different conditions with two qualitative examples.
First we infer an image which was taken with the Phenoliner in 2017 (see Fig. \ref{fig:PhenoOrig}). The illumination system and the background differ considerably from the setup in 2018. Furthermore the image shows a red variety after the veraison and features dark berries. The network is able to correctly detect most of the berries although it never saw dark berries during training (see Fig. \ref{fig:PhenoPredi}). 
As a second experiment we infer an image taken by a handheld SLR camera under natural sun light (see Fig. \ref{fig:SLROrig}). The distance between camera and canopy is different from the otherwise nearly constant 0.75 m which are recorded by the Phenoliner. The observed variety shows green berries similar to the berries from the training set. The network is able to correctly detect most of the berries as well (see Fig. \ref{fig:SLRPredi}).\\
Both examples show that our network has the potential to be used in different surroundings. 

\section{Conclusion and Outlook}
\label{sec:conclusion}
In this paper, we presented a novel berry counting approach. We are able to detect and mask single berry objects with a semantic segmentation network by using a class 'edge' to separate single objects from each other. This enables the evasion of the time and computationally intensive use of an instance segmentation network like the Mask-RCNN.
Although we trained only one network we are able to handle two training systems with different characteristics and challenges. We achieve $R^2 = 98.79~\%$ for the VSP and $97.15~\%$ for the challenging SMPH.

We compared our approach with a state-of-the-art instance segmentation approach, the Mask-RCNN. We achieve visually as well as quantitative better results ($R^2 = 81.52~\%$ for the VSP and $R^2 = 74.33~\%$ for the SMPH). Furthermore we outperform the inference time of the Mask-RCNN, which takes nearly an hour while our approach can be inferred in minutes.
The comparison with a classical regression approach yields results which are just slightly worse than our approach ($R^2 = 97.19~\%$ for the VSP and $R^2 = 92.20~\%$ for the SMPH). But the advantage of our method is the potential extraction of additional phenotypic traits like the berry size.

Despite these encouraging results, there is further space for improvements. For example, we want to investigate in detail the application to other grapevine varieties. Furthermore, we want to tackle the problem of counting berries in overlapping images multiple times. The goal is to offer the joint investigation of a whole row of plants.

There is still a gap between counting berries and estimation of yield such as 
correcting for counts in overlapping images, estimation of the number of 'invisible berries' and a proper evaluation of these steps. These steps are part of current research and beyond the scope of the paper.

\section*{Acknowledgment}
This work was funded by German Federal Ministry of Education and Research (BMBF, Bonn, Germany) in the framework of the project novisys (FKZ 031A349) 
and partially funded by the Deutsche Forschungsgemeinschaft (DFG, German Research Foundation) under Germany’s Excellence Strategy - EXC 2070 - 390732324.




\section*{References}
\bibliographystyle{model1-num-names}
\bibliography{sample}

\begin{thebibliography}{44}
\expandafter\ifx\csname natexlab\endcsname\relax\def\natexlab#1{#1}\fi
\providecommand{\bibinfo}[2]{#2}
\ifx\xfnm\relax \def\xfnm[#1]{\unskip,\space#1}\fi
\bibitem[{T\"opfer et~al.(2011)T\"opfer, Hausmann, Harst, Maul, Zyprian, and
  Eibach}]{Toepfer11}
\bibinfo{author}{R.~T\"opfer}, \bibinfo{author}{L.~Hausmann},
  \bibinfo{author}{M.~Harst}, \bibinfo{author}{E.~Maul},
  \bibinfo{author}{E.~Zyprian}, \bibinfo{author}{R.~Eibach},
  \bibinfo{title}{New Horizons for Grapvine Breeding},
  \bibinfo{publisher}{Global Science Books}, pp. \bibinfo{pages}{79--100}.
\bibitem[{Alercia et~al.(2009)Alercia, Becher, Boursiquot, Carara, amd
  A.~Costacurta, and et~al.}]{OIV01}
\bibinfo{author}{A.~Alercia}, \bibinfo{author}{R.~Becher},
  \bibinfo{author}{J.-M. Boursiquot}, \bibinfo{author}{R.~Carara},
  \bibinfo{author}{P.~C. amd A.~Costacurta}, \bibinfo{author}{et~al.},
  \bibinfo{title}{2nd edition of the oiv descriptor list for grape varieties
  and vitis species}, \bibinfo{year}{2009}.
\bibitem[{Lorenz et~al.(1995)Lorenz, Eichhorn, Bleiholder, Klose, Meier, and
  Weber}]{Lorenz95}
\bibinfo{author}{D.~Lorenz}, \bibinfo{author}{L.~Eichhorn},
  \bibinfo{author}{H.~Bleiholder}, \bibinfo{author}{R.~Klose},
  \bibinfo{author}{U.~Meier}, \bibinfo{author}{E.~Weber},
\newblock \bibinfo{title}{Growth stages of the grapevine: Phenological growth
  stages of grapevine (vitis vinifera l. ssp. vinifera) - codes and
  descriptions according to the extended bbch scale},
\newblock \bibinfo{journal}{Australian Journal of Grape and Wine Research}
  \bibinfo{volume}{1} (\bibinfo{year}{1995}) \bibinfo{pages}{133--103}.
\bibitem[{Gongal et~al.(2015)Gongal, Amatya, Karkee, Zhang, and
  Lewis}]{Gongal15}
\bibinfo{author}{A.~Gongal}, \bibinfo{author}{A.~Amatya},
  \bibinfo{author}{M.~Karkee}, \bibinfo{author}{Q.~Zhang},
  \bibinfo{author}{K.~Lewis},
\newblock \bibinfo{title}{Sensors and systems for fruit detection and
  localization: A review},
\newblock \bibinfo{journal}{Computers and Electronics in Agriculture}
  \bibinfo{volume}{116} (\bibinfo{year}{2015}) \bibinfo{pages}{8--19}.
\bibitem[{Araus and Cairns(2014)}]{Araus14}
\bibinfo{author}{J.~L. Araus}, \bibinfo{author}{J.~E. Cairns},
\newblock \bibinfo{title}{Field high-throughput phenotyping: the new crop
  breeding frontier},
\newblock \bibinfo{journal}{Trends in Plant Science} \bibinfo{volume}{19}
  (\bibinfo{year}{2014}) \bibinfo{pages}{52 -- 61}.
\bibitem[{Kipp et~al.(2014)Kipp, Mistele, Baresel, and Schmidhalter}]{Kipp14}
\bibinfo{author}{S.~Kipp}, \bibinfo{author}{B.~Mistele},
  \bibinfo{author}{P.~Baresel}, \bibinfo{author}{U.~Schmidhalter},
\newblock \bibinfo{title}{High-throughput phenotyping early plant vigour of
  winter wheat},
\newblock \bibinfo{journal}{European Journal of Agronomy} \bibinfo{volume}{52}
  (\bibinfo{year}{2014}) \bibinfo{pages}{271 -- 278}.
\bibitem[{Behmann et~al.(2015)Behmann, Mahlein, Rumpf, R\"omer, and
  Pl\"umer}]{Behmann15}
\bibinfo{author}{J.~Behmann}, \bibinfo{author}{A.-K. Mahlein},
  \bibinfo{author}{T.~Rumpf}, \bibinfo{author}{C.~R\"omer},
  \bibinfo{author}{L.~Pl\"umer},
\newblock \bibinfo{title}{A review of advanced machine learning methods for the
  detection of biotic stress in precision crop protection},
\newblock \bibinfo{journal}{Precision Agriculture} \bibinfo{volume}{16}
  (\bibinfo{year}{2015}) \bibinfo{pages}{239--260}.
\bibitem[{Foerster et~al.(2019)Foerster, Behley, Behmann, and
  Roscher}]{foerster2019}
\bibinfo{author}{A.~Foerster}, \bibinfo{author}{J.~Behley},
  \bibinfo{author}{J.~Behmann}, \bibinfo{author}{R.~Roscher},
\newblock \bibinfo{title}{Hyperspectral plant disease forecasting using
  generative adversarial networks},
\newblock in: \bibinfo{booktitle}{International Geoscience and Remote Sensing
  Symposium}.
\bibitem[{Strothmann et~al.(2019)Strothmann, Rascher, and
  Roscher}]{strothmann19}
\bibinfo{author}{L.~Strothmann}, \bibinfo{author}{U.~Rascher},
  \bibinfo{author}{R.~Roscher},
\newblock \bibinfo{title}{Detection of anomalous grapevine berries using
  all-convolutional autoencoders},
\newblock in: \bibinfo{booktitle}{International Geoscience and Remote Sensing
  Symposium}.
\bibitem[{Milioto et~al.(2018)Milioto, Lottes, and Stachniss}]{Milioto18_2}
\bibinfo{author}{A.~Milioto}, \bibinfo{author}{P.~Lottes},
  \bibinfo{author}{C.~Stachniss},
\newblock \bibinfo{title}{Real-time semantic segmentation of crop and weed for
  precision agriculture robots leveraging background knowledge in cnns},
\newblock \bibinfo{journal}{Proceedings of the IEEE Int. Conf. on Robotics \&
  Automation (ICRA)}  (\bibinfo{year}{2018}).
\bibitem[{Lottes et~al.(2019)Lottes, Behley, Chebrolu, Milioto, and
  Stachniss}]{lottes19}
\bibinfo{author}{P.~Lottes}, \bibinfo{author}{J.~Behley},
  \bibinfo{author}{N.~Chebrolu}, \bibinfo{author}{A.~Milioto},
  \bibinfo{author}{C.~Stachniss},
\newblock \bibinfo{title}{Robust joint stem detection and crop‐weed
  classification using image sequences for plant‐specific treatment in
  precision farming},
\newblock \bibinfo{journal}{Journal of Field Robotics}  (\bibinfo{year}{2019}).
\bibitem[{Roscher et~al.(2014)Roscher, Herzog, Kunkel, Kicherer, T\"opfer, and
  F\"orstner}]{Roscher13}
\bibinfo{author}{R.~Roscher}, \bibinfo{author}{K.~Herzog},
  \bibinfo{author}{A.~Kunkel}, \bibinfo{author}{A.~Kicherer},
  \bibinfo{author}{R.~T\"opfer}, \bibinfo{author}{W.~F\"orstner},
\newblock \bibinfo{title}{Automated image analysis framework for high
  throughput determination of grapevine berry size using conditional random
  fields},
\newblock \bibinfo{journal}{Computers and Electronics in Agriculture}
  \bibinfo{volume}{100} (\bibinfo{year}{2014}) \bibinfo{pages}{148--158}.
\bibitem[{Nuske et~al.(2011)Nuske, Achar, Bates, Narasimhan, and
  Singh}]{Nuske11}
\bibinfo{author}{S.~Nuske}, \bibinfo{author}{S.~Achar},
  \bibinfo{author}{T.~Bates}, \bibinfo{author}{S.~Narasimhan},
  \bibinfo{author}{S.~Singh},
\newblock \bibinfo{title}{Yield estimation in vineyards by visual grape
  detection},
\newblock \bibinfo{journal}{IEEE/RSJ International Conference on Intelligent
  Robots and Systems}  (\bibinfo{year}{2011}) \bibinfo{pages}{2352--2358}.
\bibitem[{Nyarko et~al.(2018)Nyarko, Vidovi\'{c}, Rado\u{c}aj, and
  Cupec}]{Nyarko18}
\bibinfo{author}{E.~K. Nyarko}, \bibinfo{author}{I.~Vidovi\'{c}},
  \bibinfo{author}{K.~Rado\u{c}aj}, \bibinfo{author}{R.~Cupec},
\newblock \bibinfo{title}{A nearest neighbor approach for fruit recognition in
  rgb-d images based on detection of convex surfaces},
\newblock \bibinfo{journal}{Expert Systems with Applications}
  \bibinfo{volume}{114} (\bibinfo{year}{2018}) \bibinfo{pages}{454--466}.
\bibitem[{Nuske et~al.(2014)Nuske, Wilshusen, Achar, Yoder, Narasimhan, and
  Singh}]{Nuske14}
\bibinfo{author}{S.~Nuske}, \bibinfo{author}{K.~Wilshusen},
  \bibinfo{author}{S.~Achar}, \bibinfo{author}{L.~Yoder},
  \bibinfo{author}{S.~Narasimhan}, \bibinfo{author}{S.~Singh},
\newblock \bibinfo{title}{Automated visual yield estimation in vineyards},
\newblock \bibinfo{journal}{Journal of Field Robotics} \bibinfo{volume}{31}
  (\bibinfo{year}{2014}) \bibinfo{pages}{837 -- 860}.
\bibitem[{Rose et~al.(2016)Rose, Kicherer, Wieland, Klingbeil, T\"opfer, and
  Kuhlmann}]{Rose16}
\bibinfo{author}{J.~Rose}, \bibinfo{author}{A.~Kicherer},
  \bibinfo{author}{M.~Wieland}, \bibinfo{author}{L.~Klingbeil},
  \bibinfo{author}{R.~T\"opfer}, \bibinfo{author}{H.~Kuhlmann},
\newblock \bibinfo{title}{Towards automated large-scale 3d phenotyping of
  vineyards under field conditions},
\newblock \bibinfo{journal}{Sensors} \bibinfo{volume}{16}
  (\bibinfo{year}{2016}).
\bibitem[{Rist et~al.(2018)Rist, Herzog, Mack, Richter, Steinhage, and
  T\"opfer}]{Rist18}
\bibinfo{author}{F.~Rist}, \bibinfo{author}{K.~Herzog},
  \bibinfo{author}{J.~Mack}, \bibinfo{author}{R.~Richter},
  \bibinfo{author}{V.~Steinhage}, \bibinfo{author}{R.~T\"opfer},
\newblock \bibinfo{title}{High-precision phenotyping of grape bunch
  architecture using fast 3d sensor and automation},
\newblock \bibinfo{journal}{Sensors} \bibinfo{volume}{18}
  (\bibinfo{year}{2018}) \bibinfo{pages}{763}.
\bibitem[{Krizhevsky et~al.(2012)Krizhevsky, Sutskever, and
  Hinton}]{Krizhevsky12}
\bibinfo{author}{A.~Krizhevsky}, \bibinfo{author}{I.~Sutskever},
  \bibinfo{author}{G.~E. Hinton},
\newblock \bibinfo{title}{Imagenet classification with deep convolutional
  neural networks},
\newblock \bibinfo{journal}{Advances in neural information processing systems}
  (\bibinfo{year}{2012}) \bibinfo{pages}{1097 -- 1105}.
\bibitem[{Long et~al.(2015)Long, Shelhamer, and T.Darell}]{Long15}
\bibinfo{author}{J.~Long}, \bibinfo{author}{E.~Shelhamer},
  \bibinfo{author}{T.Darell},
\newblock \bibinfo{title}{Fully convolutional networks for semantic
  segmentation},
\newblock \bibinfo{journal}{Proceedings in the IEEE Conference on Computer
  Vision and Pattern Recognition}  (\bibinfo{year}{2015}) \bibinfo{pages}{3431
  -- 3440}.
\bibitem[{He et~al.(2017)He, Gkioxari, Doll{\'a}r, and Girshick}]{He17}
\bibinfo{author}{K.~He}, \bibinfo{author}{G.~Gkioxari},
  \bibinfo{author}{P.~Doll{\'a}r}, \bibinfo{author}{R.~B. Girshick},
\newblock \bibinfo{title}{Mask r-cnn},
\newblock \bibinfo{journal}{2017 IEEE International Conference on Computer
  Vision (ICCV)}  (\bibinfo{year}{2017}) \bibinfo{pages}{2980--2988}.
\bibitem[{Lempitsky and Zisserman(2010)}]{Lemptisky10}
\bibinfo{author}{V.~Lempitsky}, \bibinfo{author}{A.~Zisserman},
\newblock \bibinfo{title}{Learning to count objects in images.},
\newblock \bibinfo{journal}{NIPS}  (\bibinfo{year}{2010})
  \bibinfo{pages}{1324--1332}.
\bibitem[{Cohen et~al.(2017)Cohen, Boucher, Glastonbury, Lo, and
  Bengio}]{Cohen17}
\bibinfo{author}{J.~P. Cohen}, \bibinfo{author}{G.~Boucher},
  \bibinfo{author}{C.~A. Glastonbury}, \bibinfo{author}{H.~Z. Lo},
  \bibinfo{author}{Y.~Bengio},
\newblock \bibinfo{title}{Count-ception: Counting by fully convolutional
  redundant counting},
\newblock \bibinfo{journal}{CoRR} \bibinfo{volume}{abs/1703.08710}
  (\bibinfo{year}{2017}).
\bibitem[{Arteta et~al.(2016)Arteta, Lempitsky, and Zisserman}]{Arteta16}
\bibinfo{author}{C.~Arteta}, \bibinfo{author}{V.~Lempitsky},
  \bibinfo{author}{A.~Zisserman},
\newblock \bibinfo{title}{Counting in the wild},
\newblock in: \bibinfo{booktitle}{European Conference on Computer Vision}.
\bibitem[{Xie et~al.(2015)Xie, Noble, and Zisserman}]{Xie16}
\bibinfo{author}{W.~Xie}, \bibinfo{author}{J.~A. Noble},
  \bibinfo{author}{A.~Zisserman},
\newblock \bibinfo{title}{Microscopy cell counting with fully convolutional
  regression networks},
\newblock \bibinfo{journal}{MICCAI 1st Workshop on Deepl Learning in Medical
  Image Analysis}  (\bibinfo{year}{2015}).
\bibitem[{Guo et~al.(2019)Guo, Stein, Wu, and Krishnamurthy}]{Guo19}
\bibinfo{author}{Y.~Guo}, \bibinfo{author}{J.~Stein}, \bibinfo{author}{G.~Wu},
  \bibinfo{author}{A.~Krishnamurthy},
\newblock \bibinfo{title}{Sau-net: A universal deep network for cell counting},
\newblock in: \bibinfo{booktitle}{Proceedings of the 10th ACM International
  Conference on Bioinformatics, Computational Biology and Health Informatics},
  BCB '19, pp. \bibinfo{pages}{299--306}.
\bibitem[{{Lobry} and {Tuia}(2019)}]{Lobry19}
\bibinfo{author}{S.~{Lobry}}, \bibinfo{author}{D.~{Tuia}},
\newblock \bibinfo{title}{Deep learning models to count buildings in
  high-resolution overhead images},
\newblock in: \bibinfo{booktitle}{2019 Joint Urban Remote Sensing Event
  (JURSE)}, pp. \bibinfo{pages}{1--4}.
\bibitem[{Lu et~al.(2018)Lu, Xie, and Zisserman}]{Lu18}
\bibinfo{author}{E.~Lu}, \bibinfo{author}{W.~Xie},
  \bibinfo{author}{A.~Zisserman},
\newblock \bibinfo{title}{Class-agnostic counting},
\newblock \bibinfo{journal}{CoRR} \bibinfo{volume}{abs/1811.00472}
  (\bibinfo{year}{2018}).
\bibitem[{{Yang} et~al.(2018){Yang}, {Yuan}, {Lunga}, {Laverdiere}, {Rose}, and
  {Bhaduri}}]{Yang18}
\bibinfo{author}{H.~L. {Yang}}, \bibinfo{author}{J.~{Yuan}},
  \bibinfo{author}{D.~{Lunga}}, \bibinfo{author}{M.~{Laverdiere}},
  \bibinfo{author}{A.~{Rose}}, \bibinfo{author}{B.~{Bhaduri}},
\newblock \bibinfo{title}{Building extraction at scale using convolutional
  neural network: Mapping of the united states},
\newblock \bibinfo{journal}{IEEE Journal of Selected Topics in Applied Earth
  Observations and Remote Sensing} \bibinfo{volume}{11} (\bibinfo{year}{2018})
  \bibinfo{pages}{2600--2614}.
\bibitem[{Marmanis et~al.(2018)Marmanis, Schindler, Wegner, Galliani, Datcu,
  and Stilla}]{Marmanis18}
\bibinfo{author}{D.~Marmanis}, \bibinfo{author}{K.~Schindler},
  \bibinfo{author}{J.~Wegner}, \bibinfo{author}{S.~Galliani},
  \bibinfo{author}{M.~Datcu}, \bibinfo{author}{U.~Stilla},
\newblock \bibinfo{title}{Classification with an edge: Improving semantic image
  segmentation with boundary detection},
\newblock \bibinfo{journal}{ISPRS Journal of Photogrammetry and Remote Sensing}
  \bibinfo{volume}{135} (\bibinfo{year}{2018}) \bibinfo{pages}{158 -- 172}.
\bibitem[{Kamilaris and Prenafeta-Boldu(2018)}]{Kamilaris2018}
\bibinfo{author}{A.~Kamilaris}, \bibinfo{author}{F.~X. Prenafeta-Boldu},
\newblock \bibinfo{title}{Deep learning in agriculture: A survey},
\newblock \bibinfo{journal}{Computers and Electronics in Agriculture}
  \bibinfo{volume}{147} (\bibinfo{year}{2018}) \bibinfo{pages}{70--90}.
\bibitem[{Aquino et~al.(2016)Aquino, Diago, Millan, and Tardaguila}]{Aquino16}
\bibinfo{author}{A.~Aquino}, \bibinfo{author}{M.~P. Diago},
  \bibinfo{author}{B.~Millan}, \bibinfo{author}{J.~Tardaguila},
\newblock \bibinfo{title}{A new methodology for estimating the grapevine-berry
  number per cluster using image analysis},
\newblock \bibinfo{journal}{Biosystem Engineering} \bibinfo{volume}{159}
  (\bibinfo{year}{2016}) \bibinfo{pages}{80 -- 95}.
\bibitem[{Aquino et~al.(2018)Aquino, Millan, Diago, and Tardaguila}]{Aquino17}
\bibinfo{author}{A.~Aquino}, \bibinfo{author}{B.~Millan},
  \bibinfo{author}{M.-P. Diago}, \bibinfo{author}{J.~Tardaguila},
\newblock \bibinfo{title}{Automated early yield prediction in vineyards from
  on-the-go image acquisition},
\newblock \bibinfo{journal}{Computers and Electronics in Agriculture}
  \bibinfo{volume}{144} (\bibinfo{year}{2018}) \bibinfo{pages}{26 -- 36}.
\bibitem[{Rudolph et~al.(2018)Rudolph, Herzog, T\"opfer, and
  Steinhage}]{Rudolph18}
\bibinfo{author}{R.~Rudolph}, \bibinfo{author}{K.~Herzog},
  \bibinfo{author}{R.~T\"opfer}, \bibinfo{author}{V.~Steinhage},
\newblock \bibinfo{title}{Efficient identification, localization and
  quantification of grapevine inflorescences in unprepared field images using
  fully convolutional networks},
\newblock \bibinfo{journal}{CoRR}  (\bibinfo{year}{2018}).
\bibitem[{Grimm et~al.(2019)Grimm, Herzog, Rist, Kicherer, T\"opfer, and
  Steinhage}]{Grimm19}
\bibinfo{author}{J.~Grimm}, \bibinfo{author}{K.~Herzog},
  \bibinfo{author}{F.~Rist}, \bibinfo{author}{A.~Kicherer},
  \bibinfo{author}{R.~T\"opfer}, \bibinfo{author}{V.~Steinhage},
\newblock \bibinfo{title}{An adaptable approach to automated visual detection
  of plant organs with applications in grapevine breeding},
\newblock \bibinfo{journal}{Biosystems Engineering} \bibinfo{volume}{183}
  (\bibinfo{year}{2019}) \bibinfo{pages}{170 -- 183}.
\bibitem[{Nellithimaru and Kantor(2019)}]{Nellithimaru19}
\bibinfo{author}{A.~K. Nellithimaru}, \bibinfo{author}{G.~A. Kantor},
\newblock \bibinfo{title}{Rols : Robust object-level slam for grape counting},
\newblock in: \bibinfo{booktitle}{The IEEE Conference on Computer Vision and
  Pattern Recognition (CVPR) Workshops}.
\bibitem[{Kicherer et~al.(2017)Kicherer, Herzog, Bendel, Klück, Backhaus,
  Wieland, Rose, Klingbeil, L\"abe, Hohl, Petry, Kuhlmann, Seiffert, and
  T\"opfer}]{Kicherer17}
\bibinfo{author}{A.~Kicherer}, \bibinfo{author}{K.~Herzog},
  \bibinfo{author}{N.~Bendel}, \bibinfo{author}{H.-C. Klück},
  \bibinfo{author}{A.~Backhaus}, \bibinfo{author}{M.~Wieland},
  \bibinfo{author}{J.~C. Rose}, \bibinfo{author}{L.~Klingbeil},
  \bibinfo{author}{T.~L\"abe}, \bibinfo{author}{C.~Hohl},
  \bibinfo{author}{W.~Petry}, \bibinfo{author}{H.~Kuhlmann},
  \bibinfo{author}{U.~Seiffert}, \bibinfo{author}{R.~T\"opfer},
\newblock \bibinfo{title}{Phenoliner: A new field phenotyping platform for
  grapevine research},
\newblock \bibinfo{journal}{Sensors}  (\bibinfo{year}{2017}).
\bibitem[{Sandler et~al.(2018)Sandler, Howard, Zhu, Zhmoginov, and
  Chen}]{Sandler18}
\bibinfo{author}{M.~Sandler}, \bibinfo{author}{A.~G. Howard},
  \bibinfo{author}{M.~Zhu}, \bibinfo{author}{A.~Zhmoginov},
  \bibinfo{author}{L.~Chen},
\newblock \bibinfo{title}{Inverted residuals and linear bottlenecks: Mobile
  networks for classification, detection and segmentation},
\newblock \bibinfo{journal}{CoRR} \bibinfo{volume}{abs/1801.04381}
  (\bibinfo{year}{2018}).
\bibitem[{Chen et~al.(2018)Chen, Zhu, Papandreou, Schroff, and Adam}]{Chen18}
\bibinfo{author}{L.~Chen}, \bibinfo{author}{Y.~Zhu},
  \bibinfo{author}{G.~Papandreou}, \bibinfo{author}{F.~Schroff},
  \bibinfo{author}{H.~Adam},
\newblock \bibinfo{title}{Encoder-decoder with atrous separable convolution for
  semantic image segmentation},
\newblock \bibinfo{journal}{CoRR} \bibinfo{volume}{abs/1802.02611}
  (\bibinfo{year}{2018}).
\bibitem[{Milioto and Stachniss(2018)}]{Milioto18}
\bibinfo{author}{A.~Milioto}, \bibinfo{author}{C.~Stachniss},
\newblock \bibinfo{title}{Bonnet: An open-source training and deployment
  framework for semantic segmentation in robotics using cnns},
\newblock \bibinfo{journal}{CoRR}  (\bibinfo{year}{2018}).
\bibitem[{Yu et~al.(2016)Yu, Jiang, Wang, Cao, and Huang}]{Yu16}
\bibinfo{author}{J.~Yu}, \bibinfo{author}{Y.~Jiang}, \bibinfo{author}{Z.~Wang},
  \bibinfo{author}{Z.~Cao}, \bibinfo{author}{T.~Huang},
\newblock \bibinfo{title}{Unitbox: An advanced object detection network},
\newblock in: \bibinfo{booktitle}{MM 2016 - Proceedings of the 2016 ACM
  Multimedia Conference}, volume \bibinfo{volume}{10072}, pp.
  \bibinfo{pages}{516--520}.
\bibitem[{Deng et~al.(2009)Deng, Dong, Socher, Li, Li, and
  Fei-Fei}]{imagenet_cvpr09}
\bibinfo{author}{J.~Deng}, \bibinfo{author}{W.~Dong},
  \bibinfo{author}{R.~Socher}, \bibinfo{author}{L.-J. Li},
  \bibinfo{author}{K.~Li}, \bibinfo{author}{L.~Fei-Fei},
\newblock \bibinfo{title}{{ImageNet: A Large-Scale Hierarchical Image
  Database}},
\newblock in: \bibinfo{booktitle}{CVPR09}.
\bibitem[{Ren et~al.(2015)Ren, He, Girshick, and Sun}]{Ren15}
\bibinfo{author}{S.~Ren}, \bibinfo{author}{K.~He},
  \bibinfo{author}{R.~Girshick}, \bibinfo{author}{J.~Sun},
\newblock \bibinfo{title}{Faster r-cnn: Towards real-time object detection with
  region proposal networks},
\newblock \bibinfo{journal}{NIPS}  (\bibinfo{year}{2015}).
\bibitem[{Abdulla(2017)}]{Abdulla17}
\bibinfo{author}{W.~Abdulla}, \bibinfo{title}{Mask r-cnn for object detection
  and instance segmentation on keras and tensorflow},
  \bibinfo{howpublished}{\url{https://github.com/matterport/Mask_RCNN}},
  \bibinfo{year}{2017}.
\bibitem[{Ronneberger et~al.(2015)Ronneberger, Fischer, and
  Brox}]{Ronneberger15}
\bibinfo{author}{O.~Ronneberger}, \bibinfo{author}{P.~Fischer},
  \bibinfo{author}{T.~Brox},
\newblock \bibinfo{title}{U-net: Convolutional networks for biomedical image
  segmentation},
\newblock \bibinfo{journal}{Medical Image Computing and Computer-Assisted
  Intervention (MICCAI)} \bibinfo{volume}{9351} (\bibinfo{year}{2015})
  \bibinfo{pages}{234--241}.

\end{thebibliography}







\end{document}